\documentclass[twocolumn,amsmath,amssymb,floatfix,nature,shownopacs,footinbib,superscriptaddress]{revtex4-1} 
\usepackage{amsmath,amssymb,natbib,bm,epsf,color}
\usepackage{lineno}
\usepackage{hyperref}
\usepackage{etoolbox}
\appto\UrlBreaks{\do\-}

\usepackage{graphicx}
\usepackage{color}
\usepackage{cuted}

\usepackage{multirow}

\usepackage{algorithm}
\usepackage{algpseudocode}

\newif\ifrevmarks
\revmarkstrue

\usepackage{soul}

\begin{document}
\title{Inferring missing locations in human mobility data using memory effects}

\author{Pablo Rosillo-Rodes}\email{prosillo@ifisc.uib-csic.es}\affiliation{Instituto de F\'{\i}sica Interdisciplinar y Sistemas Complejos IFISC (CSIC-UIB), Campus UIB, 07122 Palma de Mallorca, Spain}

\author{Ant\`onia Tugores}\affiliation{Instituto de F\'{\i}sica Interdisciplinar y Sistemas Complejos IFISC (CSIC-UIB), Campus UIB, 07122 Palma de Mallorca, Spain}

\author{David S\'anchez}\affiliation{Instituto de F\'{\i}sica Interdisciplinar y Sistemas Complejos IFISC (CSIC-UIB), Campus UIB, 07122 Palma de Mallorca, Spain}

\author{Jos\'e J. Ramasco}\affiliation{Instituto de F\'{\i}sica Interdisciplinar y Sistemas Complejos IFISC (CSIC-UIB), Campus UIB, 07122 Palma de Mallorca, Spain}

\maketitle

\textbf{Empirical datasets used to characterize human mobility are sparse in space and time, owing to technological factors such as coverage and battery, or to the erratic nature of online interactions. Missing data can bias the conclusions drawn from them, so filling the gaps in incomplete timelines is a key challenge in computational social science. Here, we propose a simple imputation method in which the informativeness of an observation decays with its bidirectional temporal distance from the missing entry, at a rate set by a single parameter per individual. On synthetic and real Twitter data, a power-law decay, selected by Bayesian model comparison, yields accurate imputations (macro-AUROC above $0.9$ overall, and above $0.8$ for the most mobile users) and matches a gap-aware continuous-time Markov chain, outperforming it among the most mobile users. The method is robust to moderate noise and generalizes across countries and European administrative regions.} 

\smallskip



\begin{figure}
\includegraphics[width=\columnwidth]{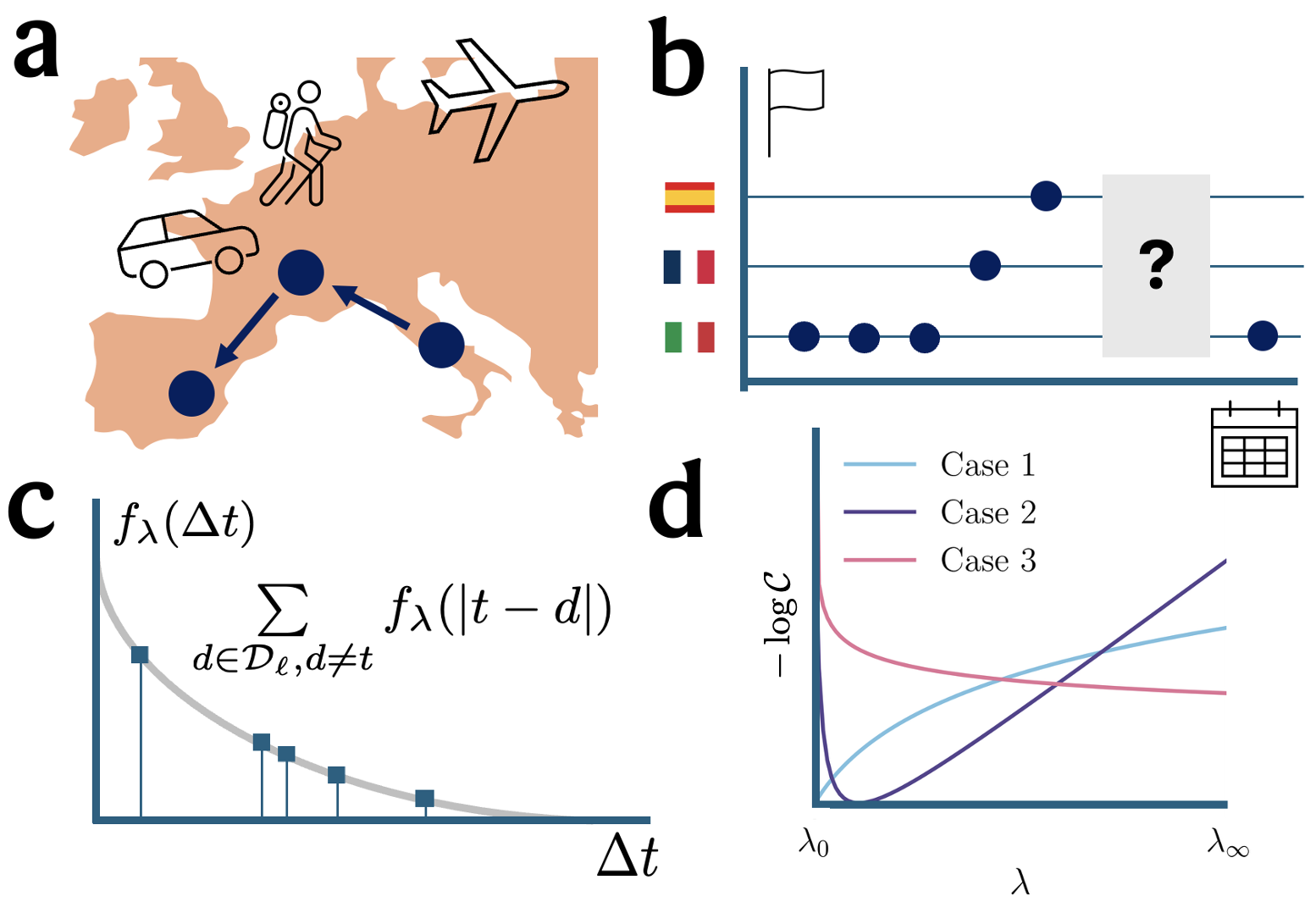}
\caption{{\bf Sketch of the method}. A human travels along different locations (\textbf{a}), which are mapped onto a timeline diagram (\textbf{b}). While certain locations are included in the mobility dataset (depicted as blue dots), others are missing (depicted as a gray rectangle or gray dots). Our goal is to infer those missing locations from a simple model. \textbf{c} We introduce a temporal weighting scheme that models the influence of past and future observations on a missing location by assigning weights based on their temporal distance $\Delta t$, according to a specified function $f_\lambda\left( \Delta t \right)$, where $\lambda$ is a decay parameter. \textbf{d} The negative log composite likelihood of the validation days, $-\log\mathcal{C}$, as a function of $\lambda$; its minimum defines $\lambda_*$. Three characteristic shapes appear. In Case~1 the minimum lies at $\lambda_0$, where the model reduces to predicting the most frequently visited location. In Case~2 the minimum is interior, and the timeline is best described by a genuine decay of informativeness with temporal separation. In Case~3 the curve decreases monotonically towards $\lambda_\infty$, where the model reduces to copying the location of the temporally nearest observed day (persistence).}
\label{fig:diapositiva1}
\end{figure}

The proliferation of mobile sensing technologies, from GPS-enabled smartphones to location-based social networks, has generated an unprecedented volume of geo-located data. Digital traces (see Fig.~\ref{fig:diapositiva1}\textbf{a}) have become fundamental for understanding human mobility \cite{Barbosa2018,Pappalardo2023}, driving applications that range from urban planning and city science \cite{louail2014,bassolas2019,dong2024} to the modeling of epidemic spreading \cite{ balcan2009, bajardi2011, barreras2024, lu2026} and flight delays~\cite{Lopez-Martin2023, Blanco2026}. These applications are of utmost importance to define public policies that depend on the analysis of scenarios produced thanks to data-driven models. The quality and potential biases of mobility data have been the focus of several recent works. Just to mention some examples, there have been discussions about the divergence of the results with different mobility data sources \cite{gallotti2024distorted}, difficulties in finding home locations with mobile phone data \cite{pappalardo2021evaluation}, biases induced by the heterogeneous activity of the users \cite{schlosser2021biases}, differences in the data between geographical areas \cite{nijs2025data} or socioeconomic biases in the representation of a population \cite{Cabrera2025ASM}. These biases have, for instance, an impact on modeling epidemic spreading and models to react against epidemics and pandemics \cite{Abrar2023AnalysisOP, barreras2024, pullano2024mobility}.

There is, however, one point that has received less attention. When considering user by user, mobility data is frequently sparse and features significant gaps or empty spaces~\cite{Barnett2016}, resulting from signal loss, device deactivation, or privacy-preserving sampling mechanisms. Consequently, the ability to accurately fill these missing data in geo-located timelines is a prerequisite for reliable behavioral analysis.  

From a brute-force perspective, deep learning predictive models have achieved high performance by learning complex, non-linear dependencies in trajectory data~\cite{Luca2021}. These methods, however, predominantly enforce spatial correlation. For instance, the Spatial Temporal Recurrent Neural Network (ST-RNN)~\cite{Liu2016} partitions the study area into grid cells and models the transition matrices between these spatial regions. While this method incorporates temporal contexts, the core architecture relies on learning spatial transitions where the likelihood of a next location is constrained by the distance from the current state. More recent attention-based models, such as DeepMove~\cite{Feng2018}, have significantly improved the capture of long-term periodicity by attending to historical trajectories. Despite their performance, these deep learning models remain limited by a heavy dependence on spatial proximity and a lack of explainability. 

Inferring missing data from a more explainable approach usually requires assumptions on statistical regularities and mechanistic models. Existing theoretical frameworks for mobility modeling and trajectory reconstruction are largely focused on methods that couple temporal dynamics with strong spatial constraints, often assuming that future locations are heavily dependent on geographical proximity to previous observations \cite{Barbosa2018}. A seminal mechanistic contribution in this domain is the Exploration and Preferential Return (EPR) model~\cite{Song2010scaling}. This model characterizes mobility as a competition between two mechanisms, namely the tendency to explore new locations (exploration) and the propensity to return to previously visited ones (preferential return). Although EPR successfully reproduces scale-free spatial patterns and the heavy-tailed distribution of jump lengths observed in empirical data~\cite{Gonzalez2008}, this model struggles to describe the long-term evolution of an individual's set of familiar locations, assuming day-to-day mobility is stationary over time. In reality, the set of familiar locations evolves over months or years~\cite{Alessandretti2018}. 

In this work, we introduce a Bayesian-inspired statistical model to impute missing data in geo-located timelines, relying exclusively on time, and learning from both past and future locations. The model takes as input only the pair (time, location label) and treats the label as an opaque categorical identifier: no coordinates, distances or adjacency relations between locations are used. This is a deliberate restriction. It delimits the setting in which no spatial information is available. 

Our approach requires that the mechanism generating the gaps is ignorable given
timing, i.e., missing at random (MAR)~\cite{Rubin1976}: whether a day is recorded does not depend on the
location occupied on that day. Additionally, we hypothesize that temporal proximity and memory effects play a key role in location inference, since observations from days closer in time to a missing entry are more informative than those further away. To test this, we design and compare several probabilistic models that assign weights to observed data based on their temporal distance from the missing entries. Our method demonstrates strong performance in inferring missing locations at both the country-level resolution and narrower geographical European unit levels in the activity of a large number of Twitter users. We put forward, thus, a parsimonious, fully interpretable mechanistic framework that integrates explainable machine learning with individual trajectory analysis leveraging a memory-decay mechanism. This mechanism characterizes how rapidly the
informativeness of an observation decays with temporal separation, at a specific, measurable rate. We use ``memory'' throughout in this statistical sense, as a property of the mobility record rather than a claim about individual cognition. 

Our main finding is the following. Since human mobility and online activity display heavy-tailed temporal statistics~\cite{Gonzalez2008,Song2010scaling,Karagiannis2007,Rybski2013}, we exploit this to recover missing entries, using both past and future information. We find that a single parameter per individual, describing how fast an observation loses informativeness, recovers most of what an explicit transition model achieves. Remarkably, our method performs on par with a gap-aware continuous-time Markov chain, which requires $K(K-1)+1$ parameters for an individual visiting $K$ locations, and outperforms it among the most mobile individuals, where imputation is hardest.

\section*{Results}

\subsection*{A memory-decay model for missing locations}
\label{sec:methods_inference}

We consider the activity of each individual as a discrete time series indexed by arbitrary time units, as illustrated in Fig.~\ref{fig:diapositiva1}\textbf{b}. We call the time units calendar days for convenience for the application we develop, but they could be other units depending on the time resolution of the mobility data. On each day, the location of individual $i$, $x^i(t)$, is either observed or missing, and the objective is to estimate the most likely location for each missing entry from the observed locations of the same individual. We denote by $\mathcal{T}_i$ the set of days on which $i$ is observed, by
$n_i = |\mathcal{T}_i|$ their number, by $\mathcal{L}$ the set of locations
present in the whole database at the geographical scale of work (countries,
provinces, cities, etc.) and by $\mathcal{L}_i \subseteq \mathcal{L}$ the
locations in which $i$ has been observed. We write the \emph{record} of
individual $i$, that is, the observed days together with the locations recorded
on them, as
\begin{equation}
\label{eq:data}
  \mathcal{D}_i = \bigl\{\,\bigl(s,\,x^i(s)\bigr) \;:\; s \in \mathcal{T}_i \,\bigr\}.
\end{equation}
We define the probability that the location of $i$ at a day
$t \notin \mathcal{T}_i$ is $\ell \in \mathcal{L}_i$ as
\begin{equation}
\label{eq:pmodel1}
    P\!\left[x^i(t) = \ell \,\middle|\, \mathcal{D}_i, \lambda\right]
    = \frac{w^i_\ell(t;\lambda)}{\sum_{k \in \mathcal{L}_i} w^i_{k}(t;\lambda)},
\end{equation}
where $w^i_\ell$ is a location-specific and time-dependent memory kernel,
\begin{equation}
\label{eq:weights}
    w^i_\ell(t;\lambda)
    = \sum_{s \in \mathcal{T}_i} f_\lambda(|t - s|)\, \delta_{\ell,x^i(s)}.
\end{equation}
Here, $\delta_{\ell,x^i(s)}$ equals one if $x^i(s) = \ell$ and zero otherwise,
and $f_\lambda(\cdot)$ is a decaying function of the temporal distance with a
single parameter $\lambda \in \mathbb{R}^+$ fitted for each individual
(Fig.~\ref{fig:diapositiva1}\textbf{c}). The most likely location at $t$ is therefore among those that $i$ has visited often at times close to $t$, both in the past and in the future of the timeline. We consider three functional forms: a power-law decay,
\begin{equation}
\label{eq:eppl}
  f_\lambda(\Delta) = \Delta^{-\lambda}, \ \ \ \mathrm{(PL)}
\end{equation}
an exponential decay,
\begin{equation}
\label{eq:epe}
  f_\lambda(\Delta) = e^{- \lambda \, \Delta }, \ \ \ \mathrm{(EXP)}
\end{equation}
and a Gaussian decay,
\begin{equation}
\label{eq:epg}
      f_\lambda(\Delta) = e^{-\lambda \, \Delta^2/2}, \ \ \ \mathrm{(GAUSS)}
\end{equation}
and, as an additional check, a stretched-exponential decay,
\begin{equation}
    \notag
f_{\lambda,\beta}(\Delta) = e^{-\lambda \Delta^{\beta}}, \ \ \ \mathrm{(STREXP)},
\end{equation}
which requires a shape parameter $\beta$ in addition to $\lambda$ (Methods).

\begin{figure*}
\centering
\includegraphics[width=\linewidth]{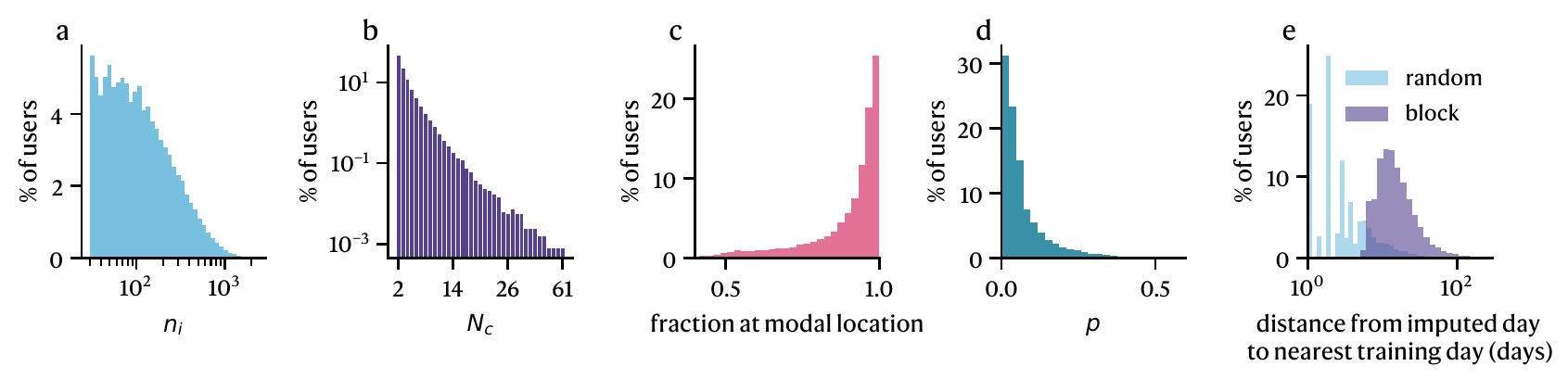}
\caption{\textbf{The empirical dataset.} Distributions over the $10^6$ country-level timelines of \textbf{a} the number of observed days $n_i$, \textbf{b} the number of distinct countries $N_c$, \textbf{c} the fraction of observed days spent at the modal country, and \textbf{d} the change rate $p$. \textbf{e} Per-individual median temporal distance from a test day to its nearest training day, under random hold-out (median over individuals $3.0$ days) and under $14$-day block hold-out ($14.0$ days).}
\label{fig:dataset}
\end{figure*}

\begin{figure*}
\includegraphics[width=\textwidth]{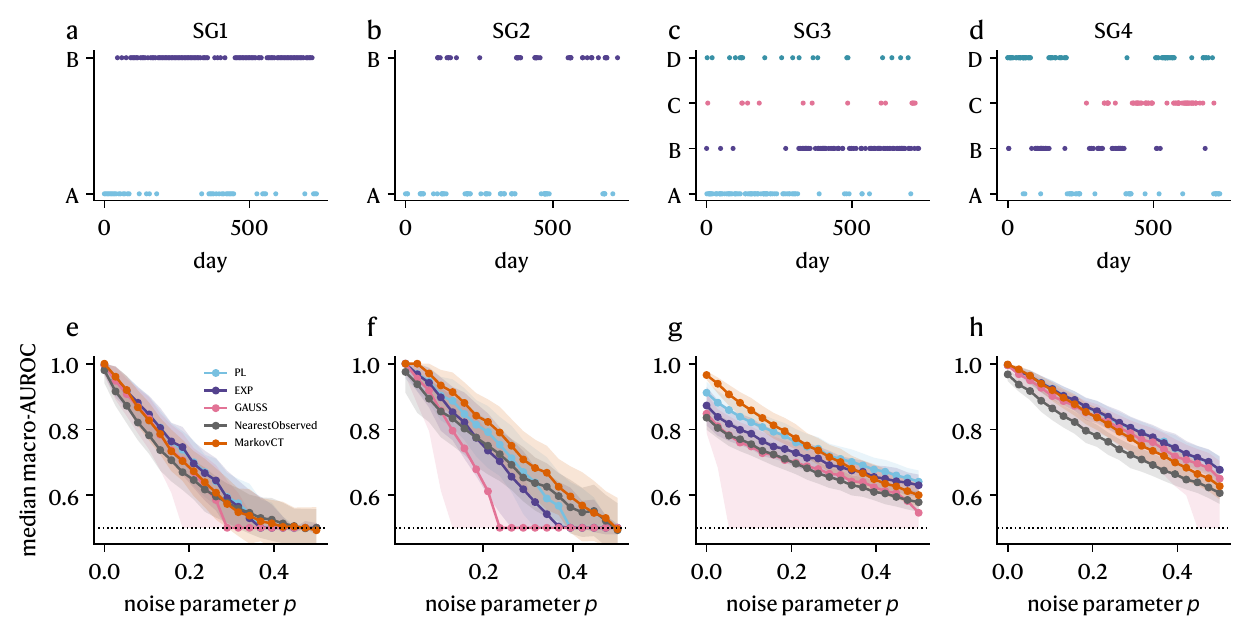}
\caption{\textbf{Synthetic generators.} \textbf{a--d} One analysed trajectory
from each generator at noise $p\approx0.13$: \textbf{a} SG1, alternation between
two locations with fixed residence intervals ($P=5$, $\Delta t_1=18$,
$\Delta t_2=54$); \textbf{b} SG2, heavy-tailed inter-event times, with a switch
of location at each observation with probability $p$; \textbf{c} SG3, four
locations with a change of home location and jittered excursions; \textbf{d}
SG4, four locations with log-normal, non-periodic residence durations. Colors
and vertical positions indicate the location. \textbf{e--h} Median test-fold
macro-AUROC as a function of the noise parameter $p$ for the three kernels, the
nearest observed day and the gap-aware Markov chain; bands span the
interquartile range over the 500 trajectories at each value of $p$, and the
dotted line marks chance level. The meaning of $p$ differs between generators
(Table~V).}
\label{fig:synthetic}
\end{figure*}

Equations~\eqref{eq:pmodel1} and~\eqref{eq:weights} are not only a smoothing construction. They follow exactly from a generative process in which each day takes its location from one observed day, chosen with probability proportional to $f_\lambda(|t-s|)$ (Methods, Eqs.~\eqref{eq:source} and~\eqref{eq:marginalization}). This interpretation identifies $\lambda$ as governing which observation is copied, and it defines the objective used to fit $\lambda$: a composite conditional likelihood over the held-out days, whose factorization across days is exact under the generative process and whose implication for real trajectories we discuss in Methods.

The two ends of the range of $\lambda$ have direct interpretations (see Fig.~\ref{fig:diapositiva1}\textbf{d} and Methods). As $\lambda \to 0$ all observations receive equal weight and Eq.~\eqref{eq:pmodel1} reduces to predicting the location the individual has visited most often, a rich-get-richer mechanism with no temporal structure~\cite{Gonzalez2008}. As $\lambda \to \infty$ the temporally nearest observation dominates, in either direction, and the model reduces to persistence, i.e., to copying the location of the closest observed day. The memory kernel therefore interpolates between two standard heuristics, which we also include among the considered baselines for validation, and an optimal parameter $\lambda_*$ in the interior of the range indicates that a gradual decay of informativeness explains the timeline better than either limit. 

To fit and evaluate the model, we divide the observed days of each individual into three disjoint subsets: a training set ($50\%$), which defines the kernel of Eq.~\eqref{eq:weights} and the candidate set $\mathcal{L}_i$; a validation set ($25\%$), used to fit $\lambda$ and to compare kernel families; and a test set ($25\%$), used to compute every performance metric reported below. Held-out days are drawn at random throughout the timeline, which makes the task interpolation within the record. Additionally, we assign held-out days in whole $14$-day calendar blocks,  in order to remove the adjacency between held-out and training days, mimicking realistic gap filling. Since we can only mask recorded days, both schemes impose a deletion mechanism that is independent of the location on the deleted day, and the resulting figures characterize performance under ignorable missingness (Methods and Discussion). To rule out any leakage, throughout the evaluation Eqs.~\eqref{eq:pmodel1}
and~\eqref{eq:weights} are computed on the training days only: the sum in
Eq.~\eqref{eq:weights} runs over $\mathcal{T}^{\text{train}}_i$ instead of
$\mathcal{T}_i$, the conditioning record is accordingly
$\mathcal{D}^{\text{train}}_i = \{(s,x^i(s)) : s \in \mathcal{T}^{\text{train}}_i\}$,
and the candidate set $\mathcal{L}_i$ is that of the locations present in the
training days. Equations~\eqref{eq:data}--\eqref{eq:weights} are stated with the
full record $\mathcal{D}_i$ because that is how the method is applied to genuine
gaps, where every observed day is available.

We compare the different kernel families through pseudo-Bayes factors, the ratio of the marginal composite likelihoods of the validation days obtained by integrating over $\lambda$ (Methods, Eqs.~\eqref{eq:bayesian_evidence} and~\eqref{eq:bf}). We perform imputations with the fitted $\lambda_*$ and with the posterior-averaged predictive (Methods, Eq.~\eqref{eq:postavg}). We compare the model with five baselines: the most frequent location, the last observed location, the nearest observed location on either side, a first-order Markov chain, and a gap-aware continuous-time Markov chain (Methods). We summarize the full procedure in Algorithm~\ref{alg:main}. We also show in Supplementary Note~4 that reinforcing preferential return by a multiplicative factor in the memory kernel~\cite{barabasi1999emergence, Gonzalez2008,Pappalardo2015, rosillo_2026} sharpens the predictions of near-stationary individuals towards their modal location but degrades those of mobile individuals, for whom imputation is non-trivial, and leaves the discrimination unchanged; we therefore do not include it. We further elaborate on these distinctions in the next section.

\subsection*{Understanding mobility patterns}

Empirical observations of real user timelines reveal a variety of mobility regimes ranging from highly regular to more erratic patterns~\cite{Song2010}. To understand these regimes and their influence on location inference, we complement the empirical data with synthetic generators that isolate the mechanisms on which the method relies, and apply the same pipeline to both real and synthetic data.

We analyze trajectories of Twitter (X) users between 2015 and 2023. We downloaded the data, formed by geo-located tweets with their time and location, through the Twitter (X) API while it remained freely open for research purposes, and anonymized. Each trajectory is a time series of daily locations, where the location recorded for a day is the country from which the majority of the individual's tweets originated (Methods). We emphasize that this defines our ground truth: the quantity we infer, and against which we validate, is the country from which the individual would have been recorded as posting, not their physical position. Also, consider that these trajectories are already sparse. The masking procedure used for training and validation (see Methods) removes days that were recorded, so our evaluation concerns a subsample of the observed record rather than of the underlying trajectory. Therefore, our masking procedure is structurally unable to speak to days on which no location was recorded at all. We find a wide range of behaviors and temporal activity levels in the empirical data, examples of which are:

\begin{itemize}
\item Users primarily based in a given country making a brief visit to another country.
\item Users consistently located in the same country throughout the timeline, with separated bursts of activity.
\item Possible residential relocations from one country to another, indicated by a sustained shift in predominant locations.
\item Users exhibiting frequent travel with a high number of visits to distinct countries.
\item Commuting-like patterns characterized by frequent back-and-forth changes in locations which correspond to nearby countries.
\end{itemize}

The country-level analysis comprises $10^6$ timelines, after removing automated accounts and applying the inclusion thresholds (at least $30$ observed days, $n_i \geq 30$, and at least two distinct locations in the training fold, $N_c \geq 2$; Methods). 
In Fig.~\ref{fig:dataset}, we show how the timelines are distributed. Timelines contain between $30$ and several thousand observed days (median $n_i = 93$, interquartile range $52$--$184$) and between $2$ and $61$ distinct countries (median $N_c = 3$). They span a median of $1258$ calendar days, of which only a median $9.4\%$ carry a location. The dataset is dominated by largely stationary individuals: the median fraction of observed days spent at the modal country is $0.944$, and the change rate $p$, the fraction of consecutive observations that imply a change of location, has median $0.042$ (interquartile range $0.020$--$0.087$). A quarter of the timelines ($25.5\%$) have $p < 0.02$, $53.3\%$ have $0.02 \le p < 0.10$, $18.2\%$ have $0.10 \le p < 0.30$, and $3.0\%$ have $p \ge 0.30$. Because a stationary individual is easy to impute for any method, we will report all results below both pooled and stratified by $p$, by $N_c$ and by $n_i$.

The sparsity of the record also determines how far held-out days lie from the observations used to impute them (Fig.~\ref{fig:dataset}\textbf{e}). Under random hold-out, the median over individuals of the median distance from a test day to the nearest training day is $3.0$ days (median over individuals of the mean distance, $5.9$ days; of the $90$th percentile, $11.1$ days). Under block hold-out these become $14.0$, $19.8$ and $36.4$ days. For a typical imputed day (see Methods), the evidence is therefore dominated by observations a few days away under random masking and two weeks away under block masking, while for roughly one day in ten it lies more than $11$ and $36$ days away, respectively. 

We now introduce synthetic data generation to evaluate our method under controlled conditions, where the generating process is known. The generators are not meant to reproduce every behavior in the dataset, but to vary separately the two ingredients that determine whether a timing-only method can work: the structure of the location process and the structure of the observation process. SG1 alternates between two locations with fixed residence intervals, observed regularly, and represents the commuting-like pattern and the extended visits to a second country; it tests whether temporal proximity recovers an alternation whose period is unknown to the method. SG2 switches between two locations with probability $p$ at each observation through heavy-tailed inter-event times, representing consistent residence with separated bursts of activity. SG3 has $K = 4$ locations and an evolving routine: the home location changes once part-way through the window, so the statistics of the early timeline do not extend to the late one, and excursions to secondary destinations occur with jittered onsets and durations, representing residential relocation, brief visits and travel to several destinations. SG4 removes periodicity altogether, with residence durations and destinations chosen at random, the irregular and non-repeating travel that none of SG1--SG3 represents. We specify the generators and the meaning of their noise parameter $p$ in Methods and Table~\ref{tab:ps}, and we show one trajectory from each in Fig.~\ref{fig:synthetic}\textbf{a}--\textbf{d} for illustration.

\subsection*{Synthetic generators}

We apply the full pipeline, with the three kernels and the five baselines, to
500 trajectories from each generator at each of 20 equally spaced values of the
noise parameter $p\in[0,0.5]$ (Methods). These values span regimes from an
almost rigid schedule to locations drawn nearly at random, so a summary pooled
over them has no clear meaning. We therefore stratify every synthetic result by
the realized change rate, using the strata of the real data (Table~I), and we
report the full dependence on $p$ in Fig.~3e--h. Because SG1, SG3 and SG4
change location even at $p=0$, only SG2 populates the stratum $p<0.02$; at
$p=0$ SG2 never switches, so its trajectories have a single location and are
excluded by the inclusion thresholds.

All methods degrade as the change rate grows, and at $p=0.5$, where each
observed day is independent of the schedule, they approach chance. The
structure of the comparison is nonetheless preserved across the range. The
memory kernels improve on the frequency and persistence heuristics in every
stratum of every generator: for $0.10\le p<0.30$ the median macro-AUROC of the
power law is 0.839, 0.800, 0.844 and 0.919 on SG1--SG4, against 0.780, 0.758,
0.772 and 0.858 for the nearest observed day, and the paired comparison favors
the power law on 83\%, 65\%, 90\% and 96\% of the trajectories in that stratum.

The gap-aware Markov chain is the strongest method on the two-location
generators, where its transition matrix is exactly the generating mechanism: it has the highest median macro-AUROC in every stratum of SG1 and SG2 except SG1 at $0.10 \leq p < 0.30$, where the power law (0.839) and the exponential kernel (0.840) are marginally ahead (0.828). On the
four-location generators the picture depends on mobility. The chain leads at
low change rates (0.970 against 0.916 on SG3 and 0.995 against 0.991 on SG4 for
$0.02\le p<0.10$), whereas the power law is ahead on SG3 for $p\ge 0.30$ (0.703
against 0.687) and on SG4 for $0.10\le p<0.30$ and $p\ge 0.30$ (0.919 against
0.912 and 0.767 against 0.726), the regime in which it also leads on the real
data. The memory model is moreover better calibrated than the chain in eight of
the thirteen strata: on SG3 at $p\ge 0.30$ its median log-loss is 1.085 against
1.527, and the paired difference favors it on 98\% of the trajectories.

The kernel comparison is stratified in the same way (Table~II). The power law is
favored over the light-tailed kernels wherever the location process has temporal
structure, and the separation grows with the number of locations: for
$0.10\le p<0.30$ the median log pseudo-Bayes factor against the exponential and
the Gaussian is 2.95 and 6.93 on SG1, 3.41 and 7.34 on SG2, 6.54 and 20.12 on
SG3, and 7.30 and 17.96 on SG4, favoring the power law on 90--98\% of the
trajectories. Against the stretched exponential the separation is an order of
magnitude smaller and vanishes in the least mobile strata. Test-fold
performance ranks the kernels in the same order, so the selection does not rest
on the prior.

\begin{table*}
\caption{\textbf{Performance on the synthetic generators.} Median test-fold
macro-AUROC of the three kernels and the five baselines, stratified by the
realized change rate $p$; $n$ is the number of trajectories in each stratum
(500 trajectories at each of 20 values of the noise parameter per generator,
random hold-out). Only SG2 populates $p<0.02$. MostFrequent assigns the same
distribution to every day, so its macro-AUROC is 0.5 by construction. The best
value in each row is in bold. The fraction of correct inferences, the log-loss
and the remaining kernels are reported in Supplementary Note~5.}
\label{tab:synthetic}
\begin{ruledtabular}
\begin{tabular}{ll r cccccccc}
Generator & Stratum & $n$ & PL & EXP & GAUSS & MostFreq. & LastObs. & NearestObs. & Markov & MarkovCT \\ \hline
\multirow{3}{*}{SG1} & $0.02\le p<0.10$ & 1103 & 0.981 & 0.980 & 0.978 & 0.500 & 0.913 & 0.944 & 0.913 & \textbf{0.993} \\
 & $0.10\le p<0.30$ & 2447 & 0.839 & \textbf{0.840} & 0.821 & 0.500 & 0.748 & 0.780 & 0.748 & 0.828 \\
 & $p\ge 0.30$ & 6450 & 0.550 & 0.541 & 0.500 & 0.500 & 0.545 & 0.556 & 0.535 & \textbf{0.557} \\
\hline
\multirow{4}{*}{SG2} & $p<0.02$ & 166 & \textbf{1.000} & \textbf{1.000} & \textbf{1.000} & 0.500 & \textbf{1.000} & \textbf{1.000} & \textbf{1.000} & \textbf{1.000} \\
 & $0.02\le p<0.10$ & 1479 & 0.962 & 0.963 & 0.948 & 0.500 & 0.889 & 0.929 & 0.889 & \textbf{0.989} \\
 & $0.10\le p<0.30$ & 3743 & 0.800 & 0.754 & 0.636 & 0.500 & 0.717 & 0.758 & 0.715 & \textbf{0.833} \\
 & $p\ge 0.30$ & 4023 & 0.500 & 0.500 & 0.500 & 0.500 & 0.550 & 0.562 & 0.531 & \textbf{0.583} \\
\hline
\multirow{3}{*}{SG3} & $0.02\le p<0.10$ & 144 & 0.916 & 0.864 & 0.833 & 0.500 & 0.797 & 0.841 & 0.876 & \textbf{0.970} \\
 & $0.10\le p<0.30$ & 3188 & 0.844 & 0.804 & 0.772 & 0.500 & 0.743 & 0.772 & 0.811 & \textbf{0.889} \\
 & $p\ge 0.30$ & 6668 & \textbf{0.703} & 0.678 & 0.640 & 0.500 & 0.620 & 0.636 & 0.643 & 0.687 \\
\hline
\multirow{3}{*}{SG4} & $0.02\le p<0.10$ & 935 & 0.991 & 0.991 & 0.987 & 0.500 & 0.918 & 0.956 & 0.936 & \textbf{0.995} \\
 & $0.10\le p<0.30$ & 3023 & \textbf{0.919} & 0.918 & 0.899 & 0.500 & 0.825 & 0.858 & 0.841 & 0.912 \\
 & $p\ge 0.30$ & 6042 & \textbf{0.767} & 0.766 & 0.745 & 0.500 & 0.667 & 0.686 & 0.677 & 0.726 \\
\end{tabular}
\end{ruledtabular}
\end{table*}

\begin{table*}
\caption{\textbf{Kernel comparison on the synthetic generators.} Median over the
trajectories in each stratum of $\log\mathrm{PsBF}^{\mathrm{PL}}_j$, the log
pseudo-Bayes factor of the power law against kernel $j$ (Eq.~\eqref{eq:bf}),
stratified by the realized change rate $p$. In parentheses, the fraction of
trajectories favoring the power law. Positive values favor the power law. The
truncated power law and the log-normal decay are reported in Supplementary
Note~2.}
\label{tab:synthetic_kernels}
\begin{ruledtabular}
\begin{tabular}{ll r ccc}
Generator & Stratum & $n$ & EXP & GAUSS & STREXP \\ \hline
\multirow{3}{*}{SG1} & $0.02\le p<0.10$ & 1103 & 1.62 (0.76) & 4.06 (0.77) & $-$0.01 (0.47) \\
 & $0.10\le p<0.30$ & 2447 & 2.95 (0.98) & 6.93 (1.00) & 0.61 (0.73) \\
 & $p\ge 0.30$ & 6450 & 3.36 (1.00) & 6.83 (1.00) & 0.95 (0.93) \\
\hline
\multirow{4}{*}{SG2} & $p<0.02$ & 166 & $-$0.01 (0.26) & $-$0.01 (0.28) & 0.00 (0.17) \\
 & $0.02\le p<0.10$ & 1479 & 1.41 (0.63) & 3.81 (0.70) & 0.00 (0.46) \\
 & $0.10\le p<0.30$ & 3743 & 3.41 (0.93) & 7.34 (0.97) & 0.87 (0.77) \\
 & $p\ge 0.30$ & 4023 & 3.44 (0.99) & 6.81 (1.00) & 0.91 (0.93) \\
\hline
\multirow{3}{*}{SG3} & $0.02\le p<0.10$ & 144 & 7.80 (0.89) & 19.15 (0.94) & 2.48 (0.81) \\
 & $0.10\le p<0.30$ & 3188 & 6.54 (0.92) & 20.12 (0.98) & 1.60 (0.72) \\
 & $p\ge 0.30$ & 6668 & 3.01 (0.89) & 11.43 (0.99) & 0.09 (0.52) \\
\hline
\multirow{3}{*}{SG4} & $0.02\le p<0.10$ & 935 & 2.15 (0.76) & 5.59 (0.87) & 0.00 (0.50) \\
 & $0.10\le p<0.30$ & 3023 & 7.30 (0.90) & 17.96 (0.97) & 1.52 (0.67) \\
 & $p\ge 0.30$ & 6042 & 3.90 (0.87) & 10.54 (0.97) & 0.37 (0.57) \\
\end{tabular}
\end{ruledtabular}
\end{table*}

\subsection*{Real trajectories}
\label{sec:res_real}

We now apply the framework to the country-level Twitter timelines. Equivalent results at the finer resolution of European NUTS3 regions~\cite{nuts} are reported in Supplementary Note~1; all methods perform worse there than at country level, and the ranking conclusions below are unchanged, except that under random hold-out the power law is better calibrated than the gap-aware Markov chain (log-loss 0.380 against 0.404).

\begin{figure*}
\centering
\includegraphics[width=\linewidth]{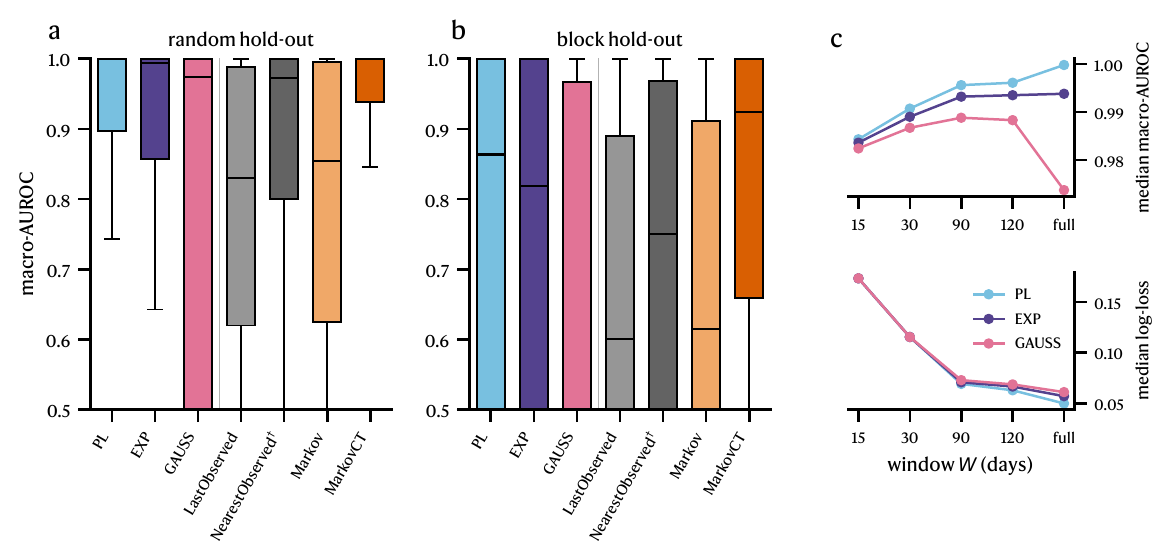}
\caption{\textbf{Comparison with baselines.} Test-fold macro-AUROC of the three memory kernels and the four baselines on the country-level timelines, \textbf{a} under random hold-out and \textbf{b} under $14$-day block hold-out. Boxes span the interquartile range across individuals and whiskers extend to $1.5$ times that range. A dagger marks the baseline that corresponds to a limiting case of the memory model; the other, MostFrequent, is omitted because, by construction, its macro-AUROC is 0.5. \textbf{c} Median macro-AUROC and log-loss when the kernel sum of Eq.~\eqref{eq:weights} is restricted to observations within $W$ days of the imputed day, for the three main kernels, and ``full'' denotes the unrestricted timeline.}
\label{fig:baselines}
\end{figure*}

\textbf{Comparison with baselines.} We compare the three kernels with the five baselines on the test folds in Table~\ref{tab:baselines} and Fig.~\ref{fig:baselines}. Because the dataset is dominated by stationary individuals, the fraction of correct inferences is uninformative in the pooled view.
We therefore focus on the macro-averaged AUROC, which weights every candidate location equally and is sensitive to the rare days away from the modal location, and on the log-loss, which measures the calibration of the predicted probabilities (Methods). Under random hold-out, the power-law model attains a median macro-AUROC of $0.9998$ and a median log-loss of $0.050$, clearly improving on the most frequent location ($0.500$, $0.194$), the last observed location ($0.830$, $0.177$), the discrete Markov chain ($0.855$, $0.139$) and the nearest observed day ($0.973$, $0.053$). Relative to the nearest observed day, which is its own $\lambda \to \infty$ limit, the median per-individual skill score of the power-law model is $0.41$ in macro-AUROC and $0.57$ in log-loss. The gap-aware Markov chain is the strongest baseline: its median macro-AUROC ($1.000$) is on par with that of the power-law model and its median log-loss is lower ($0.040$), at the cost of $K(K-1)+1$ fitted parameters per individual with $K$ locations, against a single parameter for the memory kernel. The single decay parameter thus recovers most of what the explicit transition model achieves: measured on the medians, the power-law model attains $94\%$ of the log-loss reduction of the gap-aware Markov chain relative to the most frequent location, with no transition statistics to estimate. Among the kernels, the stretched exponential performs very similarly to the power law (median macro-AUROC $1.000$ for both, median log-loss $0.054$ against $0.050$), but requires an additional shape parameter; both are better than the exponential and, especially, the Gaussian.

\begin{table*}
\caption{\textbf{Comparison with baselines on real trajectories.} Median over individuals of test-fold metrics at country level, for random hold-out and for $14$-day block hold-out. The memory kernels use the point estimate $\lambda_*$; for STREXP the shape parameter is fixed at $\beta = 0.5$. MostFrequent and NearestObserved are the $\lambda \to 0$ and $\lambda \to \infty$ limits of the memory model (Methods); MostFrequent assigns the same distribution to every day, so its macro-AUROC is $0.5$ by construction.}
\label{tab:baselines}
\footnotesize
\setlength{\tabcolsep}{3.5pt}
\begin{tabular}{l cccc c cccc}
\hline\hline
 & \multicolumn{4}{c}{Random hold-out} && \multicolumn{4}{c}{Block hold-out} \\
Method & $T$ & micro-AUROC & macro-AUROC & log-loss && $T$ & micro-AUROC & macro-AUROC & log-loss \\
\hline
PL              & 1.000 & 0.996 & 1.000 & 0.050 && 1.000 & 0.960 & 0.864 & 0.072 \\
EXP             & 1.000 & 0.994 & 0.994 & 0.057 && 1.000 & 0.957 & 0.818 & 0.072 \\
GAUSS           & 0.996 & 0.988 & 0.974 & 0.061 && 1.000 & 0.938 & 0.500 & 0.077 \\
STREXP          & 1.000 & 0.996 & 1.000 & 0.054 && 1.000 & 0.960 & 0.850 & 0.072 \\
\hline
MostFrequent    & 0.970 & 0.909 & 0.500 & 0.194 && 1.000 & 0.882 & 0.500 & 0.160 \\
LastObserved    & 0.964 & 0.950 & 0.830 & 0.177 && 0.978 & 0.889 & 0.601 & 0.138 \\
NearestObserved & 0.989 & 0.978 & 0.973 & 0.053 && 0.986 & 0.919 & 0.750 & 0.076 \\
Markov          & 0.984 & 0.975 & 0.855 & 0.139 && 1.000 & 0.933 & 0.615 & 0.146 \\
MarkovCT        & 1.000 & 0.997 & 1.000 & 0.040 && 1.000 & 0.970 & 0.924 & 0.050 \\
\hline\hline
\end{tabular}
\end{table*}

\begin{figure*}
\centering
\includegraphics[width=\linewidth]{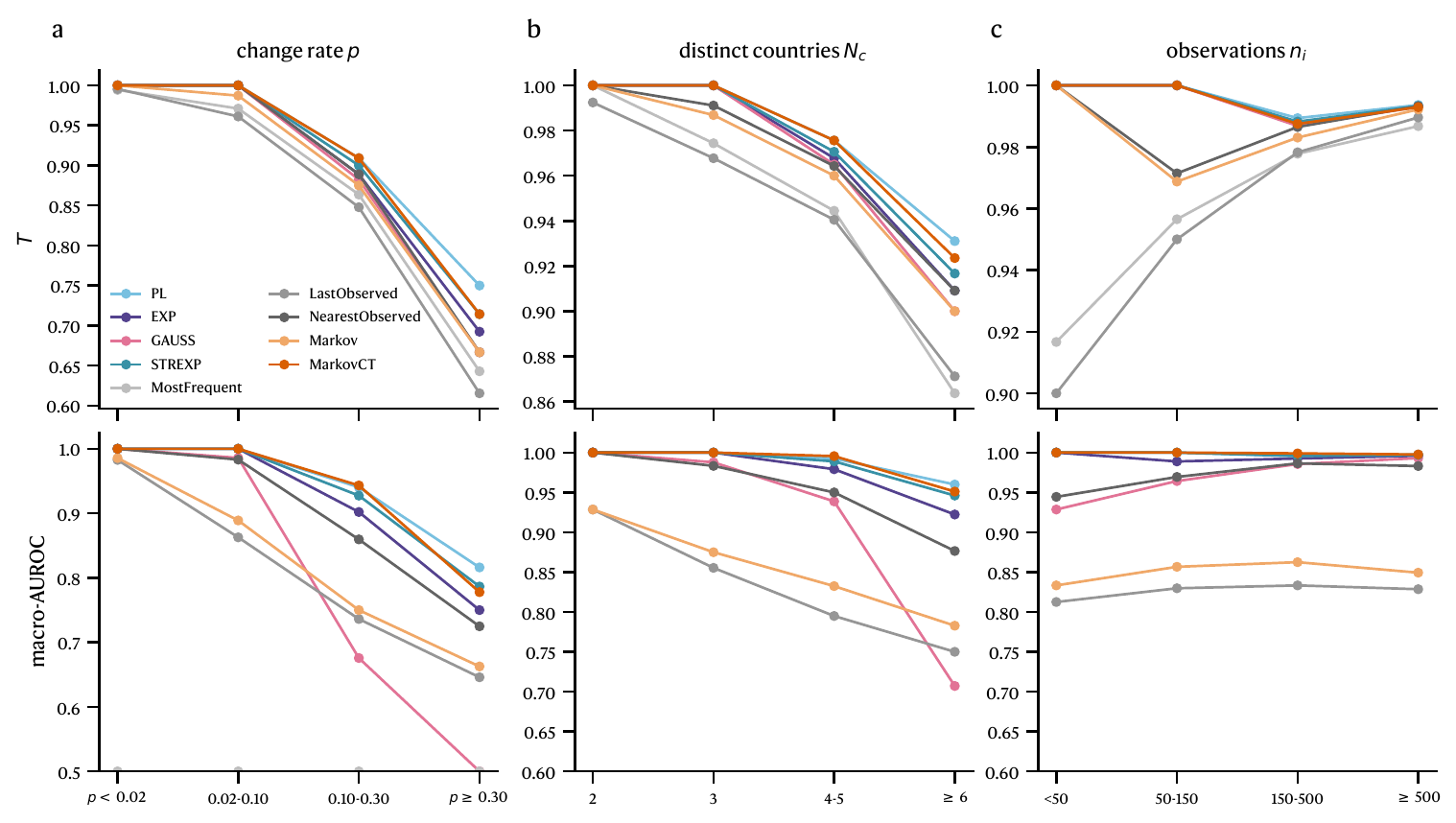}
\caption{\textbf{Performance stratified by mobility.} Median test-fold fraction of correct inferences $T$ (top) and median macro-AUROC (bottom) for selected kernels and baselines, stratified by the change rate $p$ (\textbf{a}), the number of distinct countries $N_c$ (\textbf{b}) and the number of observed days $n_i$ (\textbf{c}).}
\label{fig:stratified}
\end{figure*}

\textbf{Stratification by mobility.} In Fig.~\ref{fig:stratified} we break the comparison down by location-change rate, number of distinct countries and timeline length. For near-stationary individuals ($p < 0.02$, or $N_c = 2$) every method except the most frequent location attains a median macro-AUROC of at least $0.93$, and the fitted kernels, the nearest observed day and the gap-aware Markov chain attain $1$. The methods separate as mobility increases. For $0.10 \le p < 0.30$ the median macro-AUROC is $0.940$ for the power law, $0.902$ for the exponential, $0.676$ for the Gaussian, $0.860$ for the nearest observed day and $0.943$ for the gap-aware Markov chain; for $p \ge 0.30$ it is $0.816$, $0.750$, $0.500$, $0.725$ and $0.778$, respectively. Likewise, for individuals with $N_c \ge 6$ the power law reaches $0.960$ against $0.877$ for the nearest observed day and $0.951$ for the gap-aware Markov chain. The margin of the power-law model over the persistence baselines therefore grows with mobility, which is where the imputation problem is non-trivial, and among the most mobile individuals it also exceeds the gap-aware Markov chain. Depending on the mobility regime, a single decay parameter thus performs on par with an explicit transition model (near-stationary individuals and $0.10 \le p < 0.30$) or better than it ($p \ge 0.30$, $N_c \ge 6$). Performance depends only weakly on timeline length: the Gaussian kernel and the persistence baselines improve with $n_i$, while the power-law model stays close to its ceiling in all strata.

\begin{figure*}
\centering
\includegraphics[width=\linewidth]{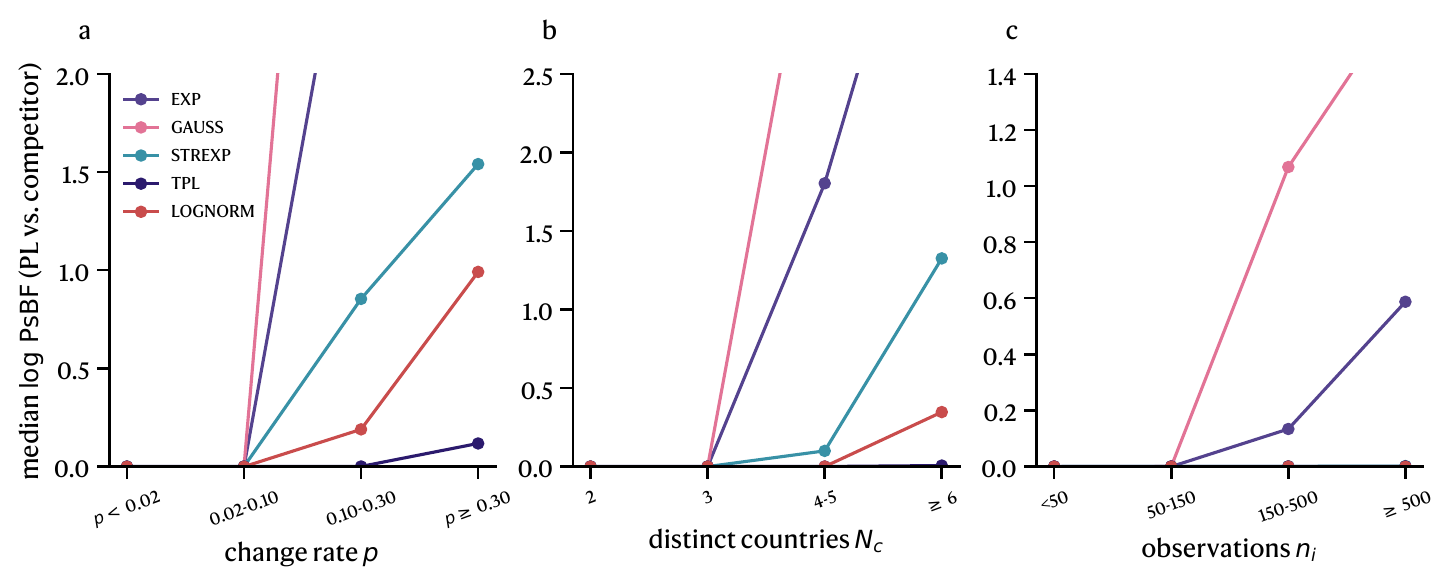}
\caption{\textbf{Kernel selection across mobility strata.} Median log pseudo-Bayes factor of the power law against each competing kernel, stratified by \textbf{a} the change rate $p$, \textbf{b} the number of distinct countries $N_c$ and \textbf{c} the number of observed days $n_i$. Positive values favor the power law. The kernels can be discriminated only among mobile individuals, where the light-tailed kernels are clearly disfavored and the stretched exponential, which requires an additional shape parameter, remains close to the power law. The truncated power law (TPL) and the log-normal decay (LOGNORM), discussed in Supplementary Note 2, are shown for reference.}
\label{fig:kernels}
\end{figure*}

\textbf{Choice of kernel.} Pooled over all individuals, the pseudo-evidence barely separates the kernels: the median log pseudo-Bayes factor between the power law and any competitor is of order $10^{-3}$, and its mean is $2.18$ against the exponential and $4.25$ against the Gaussian. The reason is the composition of the dataset. For the stationary majority every kernel predicts the modal location and the evidences coincide by construction. Stratifying by mobility shows where the functional form of the decay is identifiable (Table~\ref{tab:psbf} and Fig.~\ref{fig:kernels}). For $p \ge 0.30$ the median log pseudo-Bayes factor favoring the power law is $8.6$ against the Gaussian and $4.7$ against the exponential, and $93$--$94\%$ of these individuals favor the power law; for $0.10 \le p < 0.30$ the medians are $7.0$ and $3.3$. The same holds when stratifying by $N_c$ (Fig.~\ref{fig:kernels}\textbf{b}) and under block hold-out (Table~\ref{tab:psbf}). Against the heavy-tailed stretched exponential the separation is much smaller: the median is $1.54$ for $p \ge 0.30$ and $0.85$ for $0.10 \le p < 0.30$. This ranking does not rest on the pseudo-evidence alone, which depends on the prior over $\lambda$ (Methods): the test-fold performance at $\lambda_*$, which does not involve the prior, places the light-tailed kernels below the heavy-tailed ones in every mobile stratum (for $p \ge 0.30$ the median macro-AUROC is $0.816$ for the power law, $0.787$ for the stretched exponential, $0.750$ for the exponential and $0.500$ for the Gaussian). A preference for heavy-tailed kernels is expected, given the well-known heavy-tailed temporal statistics of human mobility~\cite{Gonzalez2008,Song2010scaling,Rybski2013}; we use the comparison to select the kernel for imputation, not as evidence for that property. The exponential and Gaussian kernels are discarded. The stretched exponential performs very similarly to the power law, in pseudo-evidence and in test-fold performance, but it needs a shape parameter $\beta$ in addition to $\lambda$, which we fixed here and which would otherwise have to be fitted for each individual. Since it brings no gain that would justify this additional parameter, we discard it in favor of the more parsimonious power law, which we use as the reference kernel throughout.

\begin{table*}
\caption{\textbf{Kernel comparison on real trajectories.} Median over individuals of $\log \mathrm{PsBF}^{\mathrm{PL}}_{j}$, the log pseudo-Bayes factor of the power law against kernel $j$ (Eq.~\eqref{eq:bf}), stratified by change rate $p$, at country level. In parentheses, the fraction of individuals favoring the power law. Positive values favor the power law.}
\label{tab:psbf}
\footnotesize
\begin{tabular}{l cccc c cc}
\hline\hline
 & \multicolumn{4}{c}{Random hold-out} && \multicolumn{2}{c}{Block hold-out} \\
$j$ & $p<0.02$ & $0.02$--$0.10$ & $0.10$--$0.30$ & $p\ge0.30$ && $0.10$--$0.30$ & $p\ge0.30$ \\
\hline
EXP     & $0.00$ (0.30) & $0.00$ (0.46) & 3.30 (0.74) & 4.73 (0.93) && 3.82 (0.74) & 4.66 (0.93) \\
GAUSS   & $0.00$ (0.32) & $0.00$ (0.47) & 6.95 (0.76) & 8.61 (0.94) && 7.12 (0.75) & 7.71 (0.94) \\
 STREXP  & $0.00$ (0.33) & $0.00$ (0.43) & 0.85 (0.67) & 1.54 (0.86) && 1.16 (0.70) & 1.54 (0.89) \\
\hline\hline
\end{tabular}
\end{table*}

\textbf{Long-range memory and gap geometry.} Two further analyses address how much of this performance depends on long-range memory and on the proximity of held-out days to training days. First, restricting the kernel sum to observations within $W$ days of the imputed day degrades the power-law model monotonically as $W$ decreases (Fig.~\ref{fig:baselines}\textbf{c}): its median macro-AUROC is $0.984$, $0.991$, $0.996$ and $0.996$ for $W = 15$, $30$, $90$ and $120$ days, against $0.9998$ for the full timeline, and its log-loss falls from $0.173$ to $0.063$ and then to $0.050$. For the exponential kernel the full timeline adds nothing beyond $W = 120$ days (macro-AUROC $0.994$ in both cases), and for the Gaussian it is detrimental ($0.974$ against $0.988$). Observations weeks or months away from a gap thus carry information, but only a heavy-tailed kernel exploits it. Part of the degradation at small $W$ comes from test days with no training observation within the window, which receive a uniform prediction; the ablation therefore measures the value of distant observations both when nearby ones exist and when they do not. Second, under block hold-out, where held-out days lie two weeks from the nearest training day, all methods lose performance but their ranking is preserved (Table~\ref{tab:baselines}): the power-law model reaches a median macro-AUROC of $0.864$ against $0.750$ for the nearest observed day, with skill scores of $0.14$ in macro-AUROC and $0.61$ in log-loss relative to it, and the gap-aware Markov chain is ahead ($0.924$) in this setting.

\textbf{Uncertainty in $\lambda$.} The fitted exponent of the power law has a median $\lambda_* = 2.1$ over the timelines with a defined AUROC, and $70\%$ of the individuals have an interior optimum. The posterior over $\lambda$ is informative only for sufficiently long timelines: its median standard deviation is $56.8$ and $55.6$ for $n_i < 50$ and $50 \le n_i < 150$, close to the value $57.7$ of the uniform prior, and it drops to $3.1$ and $1.6$ for $150 \le n_i < 500$ and $n_i \ge 500$. For short timelines, therefore, the validation fold does not determine $\lambda$ beyond the fact that it is large enough for the prediction to approach persistence. Replacing $\lambda_*$ by the posterior-averaged predictive of Eq.~\eqref{eq:postavg} leaves $T$ and the macro-AUROC essentially unchanged, reduces the pooled median log-loss from $0.050$ to $0.047$ and increases the median Brier score slightly, from $0.017$ to $0.018$. At the level of a typical individual the effect is negligible: the median per-individual improvement in log-loss is of order $10^{-5}$, marginally negative for timelines with fewer than $500$ observations and positive for longer ones, to be compared with a median per-individual gap of $0.010$ between the power-law model and the nearest observed day. We report the posterior-averaged predictive as the coherent counterpart of the marginal likelihood used for model comparison.

\textbf{Summary statistics.} Supplementary Fig.~S1 shows the joint distribution of the change rate, the micro-AUROC, $T$, the minimum negative log composite likelihood, $\lambda_*$ and $n_i$ for the power-law model over the analyzed timelines. 

\section*{Discussion}

In this work, we have introduced and evaluated a Bayesian-inspired model selection inference framework to reconstruct missing location data in user mobility timelines. By modeling the temporal influence of observed data through three different weighting schemes, namely power-law, exponential, and Gaussian, we have shown that it is possible to infer plausible mobility patterns from incomplete sequences. We formalized the inference procedure, selected the model that described the data best, proposed a likelihood-based selection of the optimal temporal decay parameter, and validated the approach on both controlled synthetic models and real Twitter trajectories extracted from geo-located tweets.

To select the kernel used for imputation, we conducted a comparative analysis using Bayes factors. These results confirm that the power-law inference framework receives the strongest support across synthetic and real datasets, capturing the underlying temporal structure of the data more effectively than the exponential or Gaussian alternatives. These are pseudo-Bayes factors, computed from a composite likelihood, and the support they give to the power law is concentrated among mobile individuals: for the stationary majority every kernel predicts the modal location and the pseudo-evidences coincide. The power law is decisively favored over the light-tailed exponential and Gaussian kernels, and test-fold performance ranks the kernels consistently. The stretched exponential performs very similarly to the power law, but it requires a shape parameter in addition to the decay rate. We therefore discard it and retain the power law, the more parsimonious of the two, as the reference kernel.

On the synthetic generators, performance degrades smoothly with the change rate
and reaches chance only where the generating process itself becomes memoryless.
Across that range the memory kernels improve on the frequency and persistence
heuristics in every stratum of every generator. The gap-aware Markov chain,
whose transition matrix is the generating mechanism of the two-location
generators, leads there in most strata and at low change rates. The memory model is ahead on
the four-location generators among the most mobile trajectories, and is better
calibrated than the chain in most strata. When applied to real-world data, the method captures broad patterns in user movement and identifies meaningful temporal scales of predictability, reflected in the distribution of optimal inference parameters and in the high values of AUROC and the fraction of correct inferences. These values, however, have to be read against the baselines and the composition of the dataset. For stationary individuals every reasonable method, including copying the nearest observed day, attains near-perfect scores, and methods separate only as mobility increases. There, the memory model clearly outperforms the frequency and persistence heuristics, which are its own limiting cases, as well as a discrete Markov chain, with a margin that grows with mobility. A gap-aware continuous-time Markov chain, with $K(K-1)+1$ parameters per individual, performs on par with it: the memory model is ahead among the most mobile individuals, whereas the Markov chain is better calibrated on average at country level and performs better when whole two-week blocks are held out. This is the central result of this work. A single parameter describing how fast an observation loses informativeness recovers most of what an explicit transition model achieves (on the medians, $94\%$ of the log-loss reduction of the gap-aware Markov chain over the most frequent location under random hold-out, and $80\%$ under block hold-out), without estimating any transition statistics, and an interpretable decay rate comes with it. 

That a heavy-tailed kernel is selected is not a new finding about human mobility, and we do not present it as one: heavy-tailed inter-event times and long-range temporal correlations in mobility and communication are well documented~\cite{Gonzalez2008,Song2010scaling,Karagiannis2007,Rybski2013}. It is rather a consistency check and a practical recommendation: the imputation works best when the kernel is compatible with this known structure, and the power law achieves this with a single parameter. We stress that the fitted exponent describes the mobility record rather than individual cognition: it may capture general temporal persistence and regularity of the record, including the burstiness of posting, and our comparison does not establish that an underlying memory process is power-law distributed.

A key strength of the method is its generality: it can be applied to any type of discrete temporal sequence with missing values, not only location data, with an appropriate adaptation of the kernels. However, this generality also implies several limitations.



First, a data-related limitation is that geo-located tweets
record a position only when a user chooses to post, so our timelines are sparse
communication records rather than continuous physical trajectories such
as GPS traces~\cite{Barnett2016}. A gap therefore signals silence, not absence,
and the observation process carries its own bursty structure~\cite{Rybski2013}
and its own selection biases, which are pervasive across mobility data
sources~\cite{nijs2025data,schlosser2021biases}. Our ground truth is thus the
country from which a user would have been recorded as posting, not their
physical position. The coarse resolution we work at makes the two largely
coincide, since posting is bursty on a timescale of days while country-level
residence changes over weeks or months; this is also why we do not claim
applicability to fine-grained trajectories. Another consequence concerns what our validation establishes. Because we can
only mask days that were recorded, the deletion mechanism we impose is unrelated
to the location on the deleted day, whereas real gaps need not have this
property: posting and movement may be coupled, and if they are, the missingness
is not ignorable. The direction of the resulting distortion is not obvious a
priori, since travel may either suppress posting through disrupted routine and
connectivity or increase it through the greater salience of posting from
unfamiliar places, and geo-located data alone cannot distinguish these. What can
be said is that a non-ignorable mechanism would have to survive aggregation to a
daily modal country and coincide with an actual change of country to bias
our inference: an individual who stops posting while traveling within their
usual country of residence introduces no error. Establishing how far this holds
would require validation against a ground-truth sensor trajectory, which we
regard as the most valuable direction for future work.


Second, our approach does not account for locations outside the set of previously observed places. An interesting avenue for future work would be to include an innovation rate to explore new locations, a feature already present in other models~\cite{Song2010scaling}.

Third, the model treats missing days as conditionally independent given the observed record. Much of the dependence between consecutive missing days is induced by the shared neighboring observations and is captured, but the residual coupling is not, so the method yields calibrated per-day probabilities rather than coherent joint reconstructions of a gap. The neglected coupling grows with the distance between an imputed day and its nearest observation, which is a few days for most imputed days in our data but exceeds two weeks for roughly one in ten under block hold-out (Fig.~\ref{fig:dataset}\textbf{e}). Inferring the copy structure of the generative process jointly across the days of a gap would remove this limitation, at the cost of a more complex sampling algorithm per individual.

Fourth, our evaluation masks recorded days either one at a time or in two-week blocks. Under random masking, the typical held-out day lies a few days from a training day, which favors any method based on temporal proximity; under block masking, all methods lose performance, the memory model less than persistence. Genuine gaps in mobility records can be longer than two weeks, and performance on them should be expected to be lower than the values we report for random masking.

Fifth, the model uses no geographical information: location labels are treated as opaque categories, so an observation at one location lends no support to a neighboring one. This is a deliberate restriction (see the Introduction), and its cost is limited at the country level. There, the candidate set $\mathcal{L}_i$ contains only the few locations at which an individual has been observed (median three countries), which partly absorbs what a distance-decay term would contribute when choosing among all possible destinations. The cost is higher at NUTS3 resolution, where candidate regions are numerous and contiguous (median six regions per individual) and distance effects should be strongest. There, the power-law model is on par with the gap-aware Markov chain under random hold-out (median macro-AUROC 0.821 against 0.817). Under 14-day block hold-out, however, all methods lose most of their discrimination, and the power law falls behind the chain (0.547 against 0.618; Supplementary Note~1). Long gaps among many contiguous regions are thus the regime in which timing-only inference is weakest. A natural extension would replace the indicator in Eq.~\eqref{eq:weights} by a spatial kernel,
\begin{equation}
\notag
    \tilde{w}^i_\ell(t;\lambda,\kappa) = \sum_{s\in\mathcal{T}^{\mathrm{train}}_i}
  f_\lambda(|t-s|)\, K_\kappa\!\big(d(\ell, x^i(s))\big),
  \ \ K_\kappa(0)=1,
  \label{eq:spatial_kernel}
\end{equation}
where $d(\ell,\ell')$ is the geographical distance between locations and $K_\kappa$ a decreasing function, for instance $K_\kappa(d)=e^{-\kappa d}$. This multiplicative factor lets an observation support nearby candidate locations. It recovers Eq.~\eqref{eq:weights} as $\kappa\to\infty$, where
$K_\kappa\!\big(d(\ell,x^i(s))\big) \to \delta_{\ell,x^i(s)}$, and it adds a
single spatial parameter that can be fitted and compared with the same
validation-fold procedure used for $\lambda$.

A comparison with deep-learning predictors~\cite{Liu2016,Feng2018} or with mechanistic models such as EPR~\cite{Song2010scaling} would be informative, but it would not be like-for-like. These approaches address next-location forecasting and draw on coordinates, spatial transitions or dense check-in histories, whereas our setting is imputation from timestamps and categorical labels alone. Our baselines were chosen to span what is available in that setting, from frequency and persistence to an explicit gap-aware transition model. Combining a timing-only memory kernel with the spatial and sequential representations learned by such models is a natural direction for hybrid approaches.

Finally, while our framework does not provide high-resolution location reconstruction, it could be used as a building block in more complex location-inference pipelines. For example, it can serve as a pre-processing step for residence detection algorithms that require continuous time series~\cite{Ruths2020}, or be used to augment mobility datasets in large-scale studies of urban dynamics, migration, or disease spread. Its simplicity (one parameter per individual, no transition statistics to estimate and no spatial information required), its interpretability and its strong empirical performance make it a valuable starting point for scalable and adaptable inference in behavioral data analysis.

\section*{Methods}

\subsection*{Data partition and evaluation geometry}

Let us make precise the partition announced in Results, and the
substitution of $\mathcal{T}_i$ by $\mathcal{T}^{\text{train}}_i$ in
Eqs.~\eqref{eq:pmodel1} and~\eqref{eq:weights} that it entails.

The observed days of each individual, $\mathcal{T}_i$, are divided into three disjoint subsets. The training set $\mathcal{T}^{\text{train}}_i$, comprising $50\%$ of the observed days, enters the definition of the kernel in Eq.~\eqref{eq:weights} and fixes the candidate location set $\mathcal{L}_i$. The validation set $\mathcal{T}^{\text{val}}_i$, comprising $25\%$, consists of days whose locations are known but treated as missing: we infer them and fit $\lambda$ so as to maximize $\mathcal{C}_i(\lambda)$ against the true positions, and we use the same days to compute the pseudo-evidence of each kernel. The test set $\mathcal{T}^{\text{test}}_i$, the remaining $25\%$, is held aside and used exactly once, to compute the performance metrics. 

We use two assignment schemes. In the random scheme, individual observed days are assigned to the three subsets uniformly at random throughout the timeline, so the task is interpolation within a completed record rather than forward extrapolation. Using observations later than a held-out day to infer it is intrinsic to this task, which is a smoothing problem that conditions on evidence on both sides of each gap, and is not a form of leakage: no held-out day enters the kernel, the candidate set or the fit of $\lambda$ for the days it is evaluated on. For the same reason, the appropriate persistence baseline is the nearest observed day on either side, and we report it together with the last observed day. In the block scheme, the calendar is divided into consecutive $14$-day blocks and whole blocks are assigned to the test and validation sets until each reaches its target fraction of observed days, so that held-out days are not adjacent to training days. For each held-out day we record the distance to the nearest training day, $\min_{s \in \mathcal{T}^{\text{train}}_i} |t-s|$, and summarize it per individual by its median, mean and $90$th percentile.

Since only days that were recorded can be masked, both schemes impose a deletion mechanism that is by construction independent of the location on the deleted day, i.e., missing completely at random with respect to location. The resulting figures therefore characterize performance under ignorable missingness. For the method to apply to genuinely missing days, the gaps in the record must be missing at random (MAR) given timing: whether a day is recorded may depend on its temporal position, but not on the location occupied on that day~\cite{Rubin1976}. We elaborate on this assumption in the Discussion.

\subsection*{Generative reading of the memory kernel}

Equations~\eqref{eq:pmodel1} and~\eqref{eq:weights} admit a generative reading, which we make explicit because it determines the status of the objects defined below. Let $c^i(t) \in \mathcal{T}^{\text{train}}_i$ be a latent source: the observed
day from which the location at $t$ is copied. We take
\begin{equation}
\label{eq:source}
  P\!\left[c^i(t) = s \mid \mathcal{D}^{\text{train}}_i, \lambda\right]
  = \frac{f_\lambda(|t-s|)}
         {\sum_{u \in \mathcal{T}^{\text{train}}_i} f_\lambda(|t-u|)}, \ \ s \in \mathcal{T}^{\text{train}}_i,
\end{equation}
so that a day is more likely to be copied the closer it lies to $t$, in either
direction, and we set $x^i(t) = x^i\!\left(c^i(t)\right)$. Marginalizing over
the source recovers Eq.~\eqref{eq:pmodel1} exactly,
\begin{align}
\notag
&P\!\left[x^i(t) = \ell \mid \mathcal{D}^{\text{train}}_i, \lambda\right] = 
   \\ &\sum_{s \in \mathcal{T}^{\text{train}}_i} \delta_{\ell,x^i(s)}\,
    \frac{f_\lambda(|t-s|)}
         {\sum_{u \in \mathcal{T}^{\text{train}}_i} f_\lambda(|t-u|)}
  = \frac{w^i_\ell(t;\lambda)}
          {\sum_{k \in \mathcal{L}_i} w^i_{k}(t;\lambda)},
  \label{eq:marginalization}
\end{align}
where the last equality follows from partitioning
$\mathcal{T}^{\text{train}}_i$ by location,
$\sum_{u \in \mathcal{T}^{\text{train}}_i} f_\lambda(|t-u|)
 = \sum_{k \in \mathcal{L}_i} w^i_{k}(t;\lambda)$.

\subsection*{Limiting cases}

The two ends of the range of $\lambda$ have a direct reading. Since a day to be
imputed is not itself a training day, temporal distances to the training days
take integer values $\Delta \ge 1$, and the four kernels we consider, those of
Eqs.~\eqref{eq:eppl}--\eqref{eq:epg} and the stretched exponential of
Eq.~\eqref{eq:strexp}, all satisfy $f_0(\Delta) = 1$. At $\lambda = 0$ every
training day therefore carries the same weight, $w^i_\ell(t;\lambda)$ reduces to
the number of training days spent at $\ell$, and Eq.~\eqref{eq:pmodel1} assigns
to each $\ell \in \mathcal{L}_i$ the fraction of training days spent at $\ell$,
the same distribution for every $t$. At the opposite end, let
$\Delta_{\min}(t) = \min_{s \in \mathcal{T}^{\text{train}}_i} |t-s|$ be the
smallest temporal distance between $t$ and a training day. For the four kernels,
$f_\lambda(\Delta)/f_\lambda(\Delta_{\min}(t)) \to 0$ as $\lambda \to \infty$
whenever $\Delta > \Delta_{\min}(t)$, so after normalization only the training
days at distance $\Delta_{\min}(t)$ retain weight, and they retain it in equal
parts: $P[x^i(t) = \ell \mid \mathcal{D}^{\text{train}}_i, \lambda]$ converges to
the distribution that places all its mass on the locations of the training days
at minimum distance from $t$, split evenly among them. These two limits are the
most-frequent-location and nearest-observed-day baselines.

\subsection*{Additional kernel}

To check whether a different heavy-tailed kernel would impute better than the power law, we compare it with the stretched exponential,
\begin{equation}
\label{eq:strexp}
 f_{\lambda,\beta} = e^{-\lambda \Delta^{\beta}}. \ \ \ \mathrm{(STREXP)}
\end{equation}
This is a two-parameter family: besides the decay rate $\lambda$, it has a shape parameter $\beta$. To keep the fitting cost identical to that of the other kernels, we fix $\beta = 0.5$ and fit $\lambda$ only; its pseudo-evidence and test-fold performance are therefore conditional on this value, and a fully specified version would require fitting $\beta$ for each individual as well. Because it performs very similarly to the power law (Results), we do not pursue it further and retain the power law, which reaches the same performance with a single parameter. It differs from the other kernels only in how it interpolates between the two limiting cases.

We restrict the comparison to kernels with this property, because a kernel whose decay survives at $\lambda = 0$ cannot reach the most-frequent-location limit on a non-negative range of $\lambda$ and would not describe the same family of models.

\subsection*{Composite likelihood and conditional independence}

Fitting $\lambda$ requires an objective, which we build on the validation days:
days whose locations we know, but which we hide from the kernel. For every
$t \in \mathcal{T}^{\text{val}}_i$ we evaluate the predictive of
Eq.~\eqref{eq:pmodel1} on the training record $\mathcal{D}^{\text{train}}_i$ and
read off the probability it assigns to the location actually recorded on that
day,
\begin{equation}
\label{eq:truth}
  y^i_t \equiv x^i(t), \qquad
  t \in \mathcal{T}^{\text{val}}_i \cup \mathcal{T}^{\text{test}}_i .
\end{equation}
For two validation days $t_1$ and $t_2$ with recorded locations $y^i_{t_1}$ and
$y^i_{t_2}$ this gives
\begin{align}
\notag
&\mathcal{C}_i[t_1,t_2](\lambda) \\ &=
  P\!\left[x^i(t_1) = y^i_{t_1} \mid \mathcal{D}^{\text{train}}_i, \lambda\right]\,
  P\!\left[x^i(t_2) = y^i_{t_2} \mid \mathcal{D}^{\text{train}}_i, \lambda\right],
\label{eq:composite}
\end{align}
and, over all validation days,
\begin{equation}
\label{eq:composite_general}
\mathcal{C}_i(\lambda) = \prod_{t \in \mathcal{T}^{\text{val}}_i}
  P\!\left[x^i(t) = y^i_t \mid \mathcal{D}^{\text{train}}_i, \lambda\right].
\end{equation}
Here, the conditioning object is the training
record $\mathcal{D}^{\text{train}}_i$, that is, the record $\mathcal{D}_i$ of
Eq.~\eqref{eq:data} restricted to $\mathcal{T}^{\text{train}}_i$. Neither the validation nor the test days enter that set,
so that no held-out day informs its own prediction. Also, $y^i_t$ is the location
recorded on the held-out day $t$, meaning that the validation days are observed days that we
treat as missing. The
three subsets therefore play distinct roles: $\mathcal{T}^{\text{train}}_i$
supplies the \emph{inputs} of the predictive distribution through
$\mathcal{D}^{\text{train}}_i$, whereas $\mathcal{T}^{\text{val}}_i$ and
$\mathcal{T}^{\text{test}}_i$ supply the \emph{outcomes} against which it is
scored, the former to fit $\lambda$ and compare kernels, the latter for the
metrics we report. Also, note that $\lambda$ appears to the right of the bar because, with
the data held fixed, $\mathcal{C}_i$ is a function of $\lambda$, that is, a
likelihood: it is what we maximize to obtain $\lambda_*$ and what we integrate
against a prior to obtain the pseudo-evidence of
Eq.~\eqref{eq:bayesian_evidence}.

Equation~\eqref{eq:composite_general} is the conditional likelihood of the
process defined by Eq.~\eqref{eq:source}, and the goal is to maximize it over
$\lambda$ (see below). Since the sources $c^i(t)$ are drawn independently across $t$, missing entries are conditionally independent given the observed record, and the factorization is exact. Real trajectories need not satisfy this: the locations of consecutive missing days are correlated beyond what the observed record implies. Relative to a process with dependent missing entries, Eq.~\eqref{eq:composite_general} is a composite conditional likelihood~\cite{Besag1975,VarinReidFirth2011}. Much of the relevant dependence is nonetheless captured, since adjacent missing days share nearly the same set of neighboring observations and receive nearly identical predictive distributions. The factorization returns a probability for each missing day separately, and so cannot describe a whole gap as a single coherent sequence. We accept this because it keeps the cost of inference linear in the number of missing days, which is what makes the method usable at this scale.

This factorization has two consequences. First, the per-day performance metrics (AUROC, $T$, log-loss) are computed from the predictive of Eq.~\eqref{eq:pmodel1} and do not involve Eq.~\eqref{eq:composite_general}. The independence assumption enters only through the selection of $\lambda$ and through the pseudo-evidence. Composite-likelihood estimators remain consistent for parameters that index correctly specified marginals~\cite{VarinReidFirth2011}, so, to the extent that Eq.~\eqref{eq:pmodel1} describes the per-day marginal, $\lambda_*$ is not biased by the neglected dependence. A joint, block-level metric would return lower values than a per-day one, but this reflects the greater difficulty of the joint criterion: even an exactly calibrated independent model with per-day accuracy $T$ scores $T^n$ on a block of $n$ days. Second, the magnitude of the neglected coupling grows with the distance between an imputed day and the observations that inform it, which is why we report that distance (Fig.~\ref{fig:dataset}\textbf{e}) and evaluate under block hold-out. The natural extension follows from the generative reading: inferring the copy structure $\{c^i(t)\}$ jointly, rather than marginalizing each source independently, would couple the missing days and would remove the factorization, at the cost of a sampling algorithm per individual.

\subsection*{Fitting $\lambda$ and comparing kernels}

For each individual and kernel we evaluate $-\log \mathcal{C}_i(\lambda)$ on a grid of $10^3 + 2$ values: the exact boundaries $\lambda_0 = 0$ and $\lambda_\infty = 200$ and $10^3$ logarithmically spaced interior points between $10^{-3}$ and $0.999\,\lambda_\infty$. The logarithmic spacing resolves small exponents, and the exact boundaries make the interior or boundary status of the optimum well defined. Individual log-probabilities inside $\mathcal{C}_i$ are floored at $-100$, so that a single held-out day at a location with vanishing predicted probability does not make the objective infinite. The optimum is $\lambda_* = \arg\max_\lambda \mathcal{C}_i(\lambda)$, and three situations arise (Fig.~\ref{fig:diapositiva1}\textbf{d}):
\begin{itemize}
    \item Case 1: the negative log composite likelihood has its minimum at $\lambda_0$. As shown in Methods, Limiting cases, the validation days are then best explained by the frequency of visits, with no temporal structure, which is related to the preferential return mechanism observed in the literature~\cite{Gonzalez2008,Song2010scaling} (see Supplementary Note 4);
    \item Case 2: the minimum is interior, $\lambda_0 < \lambda_* < \lambda_\infty$. The data support a gradual decay of informativeness with temporal separation, at a rate that characterizes the individual, which covers a wide range of mobility memories reported in the literature~\cite{Pappalardo2015};
    \item Case 3: the curve decreases monotonically and $\lambda_* = \lambda_\infty$.  As shown in Methods, Limiting cases, the validation days are then best explained by persistence, i.e., by copying the location of the nearest observed day.
\end{itemize}

To compare kernel families we define, for each individual $i$ and model $\mathcal{M}$, the marginal composite likelihood of the validation days,
\begin{equation}
\label{eq:bayesian_evidence}
Z_i^{\mathcal{M}} = \int_{\lambda_0}^{\lambda_\infty} \mathcal{C}_i^{\mathcal{M}}(\lambda) \, p(\lambda)\, \mathrm{d}\lambda,
\end{equation}
with a uniform prior $p(\lambda) = 1/(\lambda_\infty - \lambda_0)$, evaluated by trapezoidal quadrature on the grid and in log space to avoid numerical underflow. Two models are compared through
\begin{equation}
\label{eq:bf}
\log \mathrm{PsBF}^{\mathcal{M}_1}_{\mathcal{M}_2, i} = \log Z_i^{\mathcal{M}_1} - \log Z_i^{\mathcal{M}_2},
\end{equation}
which is positive when the validation days of individual $i$ favor $\mathcal{M}_1$. Because $\mathcal{C}_i$ is a composite rather than a full likelihood, Eq.~\eqref{eq:bf} is a pseudo-Bayes factor in the sense of Geisser and Eddy~\cite{GeisserEddy1979}: a composite likelihood that neglects positive dependence overstates the information in the data~\cite{VarinReidFirth2011}. We therefore summarize it across individuals by its median and by the fraction of individuals with a positive value, within strata of mobility, and we corroborate every conclusion drawn from it with test-fold predictive performance, which does not depend on the prior.

The normalized posterior
$p(\lambda \mid \mathcal{D}^{\text{val}}_i) \propto \mathcal{C}_i(\lambda)\, p(\lambda)$
is available on the same grid at no extra cost. Besides imputing with the point
estimate $\lambda_*$, we compute the posterior-averaged predictive
\begin{align}
\notag
&\bar{P}\!\left[x^i(t) = \ell \mid \mathcal{D}^{\text{train}}_i,
                \mathcal{D}^{\text{val}}_i\right] = \\
&\qquad \int P\!\left[x^i(t) = \ell \mid \mathcal{D}^{\text{train}}_i, \lambda\right]\,
          p(\lambda \mid \mathcal{D}^{\text{val}}_i)\, \mathrm{d}\lambda ,
\label{eq:postavg}
\end{align}
restricted to grid points with posterior mass above $10^{-4}$ and renormalized, and we record the posterior mean and standard deviation of $\lambda$ for every individual.

\subsection*{Baselines}

All baselines are trained on the same training days and evaluated on the same test days as the memory model, and all return a probability distribution over $\mathcal{L}_i$, so that the metrics are computed identically for every method.

\textit{MostFrequent} assigns to every day the training frequencies of the locations with add-one smoothing. It is the $\lambda \to 0$ limit of the memory model.

\textit{LastObserved} places mass $1-\epsilon$ on the location of the most recent training day strictly before $t$ and spreads $\epsilon = 0.02$ uniformly, falling back on the smoothed frequencies when there is no earlier training day. It is the causal persistence baseline.

\textit{NearestObserved} does the same with the training day closest to $t$ in either direction, splitting the mass evenly in case of ties. It is the $\lambda \to \infty$ limit of the memory model and the appropriate persistence baseline for imputation.

\textit{Markov} is a first-order Markov chain on the sequence of training observations, with transition counts smoothed by a Dirichlet prior of concentration $\alpha = 1$, applied once from the last training observation before $t$ irrespective of the length of the gap.

\textit{MarkovCT} is a gap-aware version for irregular sampling: with the same transition matrix $A$, the transition probability over a gap of $\Delta$ days is $A^{\Delta/\bar{\Delta}}$, where $\bar{\Delta}$ is the mean interval between consecutive training observations, and the distributions propagated from the preceding and the following training observations are combined by a normalized product. MarkovCT is a hidden Markov model whose hidden states coincide with the observed locations; it has $K(K-1)+1$ free parameters for an individual with $K$ candidate locations, against one for the memory model, and couples consecutive days by construction, which makes it a direct reference for what explicit dependence adds on these data.

We do not include deep-learning predictors such as ST-RNN~\cite{Liu2016} or DeepMove~\cite{Feng2018}, nor the EPR model~\cite{Song2010scaling}, as baselines. These methods address next-location forecasting and rely on coordinates, spatial transitions or dense check-in histories, which are absent by design from the categorical, timing-only setting studied here (see Discussion).

\subsection*{Performance metrics and stratification}

For each individual we compute, on the test days: the fraction of correct inferences $T$, defined as the fraction of days whose most probable location is the true one; the one-vs-rest AUROC~\cite{Fawcett2006}, both micro-averaged (pooling all location--day pairs) and macro-averaged (averaging the per-location AUROC over the locations that occur and do not occur on some test day); the balanced accuracy; the log-loss,
$-\frac{1}{|\mathcal{T}^{\text{test}}_i|}
  \sum_{t \in \mathcal{T}^{\text{test}}_i}
  \log P\!\left[x^i(t) = y^i_t \mid \mathcal{D}^{\text{train}}_i, \lambda_*\right]$; and the Brier score. Predicted probabilities are clipped at $10^{-12}$ and renormalized before scoring. Macro-averaging weights every candidate location equally, and is therefore sensitive to the rare days away from the modal location that micro-averaging dilutes for largely stationary individuals. To express improvements relative to a reference method, we use the skill score $\mathrm{SS} = (m - m_{\text{ref}})/(m_{\text{perfect}} - m_{\text{ref}})$, computed per individual, with NearestObserved as the reference, $m_{\text{perfect}} = 1$ for AUROC and $0$ for log-loss.

We stratify individuals by three covariates computed on their full observed timeline, none of which is used for fitting: the change rate $p$, the fraction of consecutive observations that imply a change of location; the number of distinct locations $N_c$; and the number of observed days $n_i$. The strata are $p < 0.02$, $0.02 \le p < 0.10$, $0.10 \le p < 0.30$, $p \ge 0.30$; $N_c = 2$, $3$, $4$--$5$, $\ge 6$; and $n_i < 50$, $50$--$150$, $150$--$500$, $\ge 500$.

For the window ablation, the sum in Eq.~\eqref{eq:weights} is restricted to training days with $|t - s| \le W$, for $W \in \{15, 30, 90, 120\}$ days, and $\lambda$ is refitted for each $W$. A held-out day with no training observation within the window receives a uniform prediction.

\subsection*{Synthetic generators}

All generators produce timelines over $730$ days, with at most one observation per day.

\textit{SG1: commuting or seasonal movement with fixed residence intervals.} The user alternates between two locations $A$ and $B$ in a deterministic cyclical pattern. They are observed every $P = 5$ days, during a stay of $\Delta t_1 = 18$ observations ($90$ days) at $A$ followed by a stay of $\Delta t_2 = 54$ observations ($270$ days) at $B$, repeating. Each observed day nominally assigned to one location is replaced by the other with probability $p$. When $p = 0$ the trajectory consists of rigid location blocks; as $p$ increases up to $0.5$, it incorporates increasing short-term deviations from the cyclic schedule. Values of $p > 0.5$ would simply invert the roles of $A$ and $B$.

\textit{SG2: bursts of activity separated by variable inactivity periods.} The user alternates between two locations with inter-event times drawn from a heavy-tailed distribution, a hallmark of human activity patterns~\cite{Karagiannis2007, Rybski2013}: $\tau = \lfloor 0.2\,X + 1 \rfloor$ days, with $X$ drawn from a Lomax (Pareto type II) distribution of shape $0.25$, whose density decays as $X^{-1.25}$, and values above $100$ days are rejected. The initial location is chosen at random and, at each observation, the user switches location with probability $p$, independently of the waiting times. 

\textit{SG3: several locations with an evolving routine.} There are $K = 4$ locations. Observation times advance by $\max(1, \xi)$ days with $\xi$ exponentially distributed with mean $3$ days. The home location is $A$ until a switching day drawn uniformly between $35\%$ and $65\%$ of the window, and $B$ afterwards, so that the statistics of the early timeline do not extend to the late one. Excursions start every $U(80, 120)$ days and last $U(7, 30)$ days; during an excursion, each observed day is assigned to one of the $K-2$ secondary locations at random. Finally, with probability $p$ each observed day is replaced by a location drawn uniformly among the $K$.

\textit{SG4: non-periodic shifts.} There are $K = 4$ locations and observation times as in SG3. Residence durations are drawn from a log-normal distribution with median $45$ days and log-standard deviation $0.6$, with a minimum of $3$ days, and at the end of each residence the next location is drawn uniformly among the other $K-1$, so that nothing repeats on a schedule. Noise is added as in SG3.

For each generator we draw $n_{\mathrm{rep}}=500$ trajectories at each of
$n_p=20$ equally spaced values of $p\in[0,0.5]$, with deterministic seeds, and
apply exactly the same pipeline, kernels, baselines, metrics and inclusion
thresholds as to the real data, under random hold-out. At $p=0$ the SG2
trajectories have a single location and are excluded, so 9500 SG2 trajectories
enter the analysis against $10,000$ for each of the other generators. Because
$p$ has a different meaning in each generator (Table~\ref{tab:ps}), we stratify all
synthetic results by the realized change rate, defined on the trajectory as for
the real data, and report performance as a function of $p$ separately for each
generator.

\begin{table}[h]
\caption{Meanings of the noise parameter $p$ across generators and in the real data.}
\label{tab:ps}
\begin{tabular}{@{} l p{6.2cm} @{}}
$p$       & Meaning \\ \hline \hline
SG1       & Chance that any individual day nominally assigned to a location is replaced by the other. \\ \hline
SG2       & Probability that the user switches location relative to the previous observation. \\ \hline
SG3--4    & Chance that any observed day is replaced by a location drawn uniformly among the $K$ locations. \\ \hline
Real data & Fraction of consecutive observations that imply a change of location. \\ \hline \hline
\end{tabular}
\end{table}

\subsection*{Empirical data}

The data consist of geo-located tweets posted between 2015 and 2023, downloaded through the Twitter (X) API while it remained freely open for research purposes, and anonymized. Locations can be recorded at different resolutions, from coordinates to places. At the country level, we use the country of emission of every tweet. For NUTS3 regions~\cite{nuts} we project each Twitter place onto the NUTS3 region containing most of its area. Tweets may originate from real users, automated accounts or high-volume bots. To focus on real users, we remove trajectories showing travel speeds above $1000$~km/h (roughly the speed of a commercial airplane), trajectories with tweeting rates exceeding $0.8$ tweets per hour per day, which may indicate spam-like behavior, and trajectories whose tweeting rates are exact integers, which are likely generated by automated accounts. Each trajectory is then collapsed to one location per day, the location from which the majority of the individual's tweets originated that day, with ties broken by the last occurrence. At country (NUTS3) level, $10^6$ ($10^3$) timelines enter the pipeline.

\begin{figure*}[t]
\hrule \vspace{0.4em}
\refstepcounter{algorithm}
\noindent\textbf{Algorithm \thealgorithm:} Memory-decay imputation of a single timeline
\label{alg:main}%
\vspace{0.4em} \hrule \vspace{0.4em}
\begin{algorithmic}[1]
\Require observed timeline $\{(t, x^i(t))\}_{t \in \mathcal{T}_i}$ with $n_i \ge 30$
         and at least two distinct locations; kernel families $\{f^{M}_\lambda\}$;
         grid $\{\lambda_k\}$ ($10^3 + 2$ points: $\lambda_0 = 0$, $\lambda_\infty = 200$,
         and $10^3$ log-spaced points in $[10^{-3}, 0.999\,\lambda_\infty]$)
\State partition $\mathcal{T}_i$ into
       $\mathcal{T}^{\mathrm{train}}_i$ ($50\%$), $\mathcal{T}^{\mathrm{val}}_i$ ($25\%$),
       $\mathcal{T}^{\mathrm{test}}_i$ ($25\%$), at random or by $14$-day blocks
\State $\mathcal{L}_i \gets \{x^i(s) : s \in \mathcal{T}^{\mathrm{train}}_i\}$
       \Comment{training support only}
\State remove from $\mathcal{T}^{\mathrm{val}}_i$ and $\mathcal{T}^{\mathrm{test}}_i$
       the days with $x^i(t) \notin \mathcal{L}_i$
\If{$|\mathcal{T}^{\mathrm{train}}_i| < 8$ \textbf{or} $|\mathcal{L}_i| < 2$ \textbf{or}
    $|\mathcal{T}^{\mathrm{val}}_i| < 2$ \textbf{or} $|\mathcal{T}^{\mathrm{test}}_i| < 2$}
  \State \textbf{exclude} the timeline from this scheme
\EndIf
\For{each kernel family $M$}
  \For{each $\lambda_k$ in the grid}
    \For{each $t \in \mathcal{T}^{\mathrm{val}}_i$ and $\ell \in \mathcal{L}_i$}
           \State $w^i_\ell(t;\lambda_k) \gets \sum_{s \in \mathcal{T}^{\mathrm{train}}_i}
             f^{M}_{\lambda_k}(|t-s|)\, \delta_{\ell, x^i(s)}$
             \Comment{Eq.~\eqref{eq:weights} on $\mathcal{D}^{\mathrm{train}}_i$}
        \State $P^i_{\lambda_k}(\ell \mid t) \gets
             w^i_\ell(t;\lambda_k) \big/
             \sum_{\ell' \in \mathcal{L}_i} w^i_{\ell'}(t;\lambda_k)$
    \EndFor
        \State $\log \mathcal{C}^{M}_i(\lambda_k) \gets
           \sum_{t \in \mathcal{T}^{\mathrm{val}}_i}
           \max\bigl\{\log P^i_{\lambda_k}\bigl(y^i_t \mid t\bigr),\, -100\bigr\}$
  \EndFor
  \State $\lambda_* \gets \lambda_{k_*}$, \ $k_* = \arg\max_k \log \mathcal{C}^{M}_i(\lambda_k)$
  \State $\log Z^{M}_i \gets \log \int_{\lambda_0}^{\lambda_\infty} \mathcal{C}^{M}_i(\lambda)\, p(\lambda)\, d\lambda$
         \Comment{Eq.~\eqref{eq:bayesian_evidence}, trapezoidal, in log space}
    \State $p(\lambda \mid \mathcal{D}^{\mathrm{val}}_i) \gets
          \mathcal{C}^{M}_i(\lambda)\,p(\lambda) / Z^{M}_i$ 
  \State \textbf{predict} on $\mathcal{T}^{\mathrm{test}}_i$ at $\lambda_*$ and with Eq.~\eqref{eq:postavg}
\EndFor
\State \textbf{baselines:} fit on $\mathcal{T}^{\mathrm{train}}_i$, predict on $\mathcal{T}^{\mathrm{test}}_i$
\State \textbf{model comparison:} $\log \mathrm{PsBF}^{M_1}_{M_2,i} = \log Z^{M_1}_i - \log Z^{M_2}_i$
       \Comment{Eq.~\eqref{eq:bf}}
\State \Return imputed distributions, $\lambda_*$, posterior summaries, $\log Z^{M}_i$, test metrics
\end{algorithmic}
\vspace{0.4em} \hrule
\end{figure*}

\section*{Data and code availability statement}

The timelines of real Twitter users cannot be made available due to data protection regulations. We provide the code and the implementation of our framework in a public repository, available at Ref.~\cite{bayloc}.

\section*{Acknowledgements}

We thank Lucas Lacasa, Juan De Gregorio, Miquel Artigues, and Felipe Olivares for valuable comments and references. We also thank IFISC Computing Services Unit team for developing and maintaining the data infrastructure at IFISC. This work was partially supported by the Spanish State Research Agency
(MICIU/AEI/10.13039/501100011033) and FEDER (UE) under project COSASTI (PID2024-157493NB-C21 \& PID2024-157493NB-C22), and the Mar{\'\i}a de Maeztu project CEX2021-001164-M, and by the Government of the Balearic Islands through an
Ajut per a projectes de recerca científica i tecnològica (PRD2025\_092)

\bibliographystyle{naturemag}
\bibliography{bibliography}

@article{Besag1975,
  author  = {Besag, Julian},
  title   = {Statistical Analysis of Non-Lattice Data},
  journal = {The Statistician},
  year    = {1975},
  volume  = {24},
  number  = {3},
  pages   = {179--195},
  doi     = {10.2307/2987782},
}

@article{GeisserEddy1979,
  author  = {Geisser, Seymour and Eddy, William F.},
  title   = {A Predictive Approach to Model Selection},
  journal = {Journal of the American Statistical Association},
  year    = {1979},
  volume  = {74},
  number  = {365},
  pages   = {153--160},
  doi     = {10.1080/01621459.1979.10481632},
}

@article{VarinReidFirth2011,
  author  = {Varin, Cristiano and Reid, Nancy and Firth, David},
  title   = {An Overview of Composite Likelihood Methods},
  journal = {Statistica Sinica},
  year    = {2011},
  volume  = {21},
  number  = {1},
  pages   = {5--42},
}

@article{Lopez-Martin2023,
	author = {L{\'o}pez-Mart{\'\i}n, Ra{\'u}l and Zanin, Massimiliano},
	journal = {Aerospace},
	number = {3},
	title = {Propagation of Interactions among Aircraft Trajectories: A Complex Network Approach},
	volume = {10},
	year = {2023}}

@article{Blanco2026,
	author = {Josu Blanco and Ant{\`o}nia Tugores and Jos{\'e} J. Ramasco and Massimiliano Zanin},
	journal = {Transportation Research Part E: Logistics and Transportation Review},
	pages = {104688},
	title = {The structure of global delay propagation in air transport},
	volume = {208},
	year = {2026}}

@article{rosillo_2026,
  title={{Simon's model does not produce Zipf's law: The fundamental rich-get-richer mechanism for any power-law size ranking}},
  author={{P. Rosillo-Rodes, J. W. Zimmerman, L. Hébert-Dufresne and P. S. Dodds}},
  journal={arXiv preprint arXiv:2604.13184},
  year={2026}
}

@article{barabasi1999emergence,
  title={Emergence of scaling in random networks},
  author={Barab{\'a}si, Albert-L{\'a}szl{\'o} and Albert, R{\'e}ka},
  journal={Science},
  volume={286},
  number={5439},
  pages={509--512},
  year={1999},
  publisher={American Association for the Advancement of Science}
}

@article{Rubin1976,
  author  = {Rubin, Donald B.},
  title   = {Inference and Missing Data},
  journal = {Biometrika},
  year    = {1976},
  volume  = {63},
  number  = {3},
  pages   = {581--592},
  doi     = {10.1093/biomet/63.3.581},
}

@article{Song2010,
  author       = {Chaoming Song and Zehui Qu and Nicholas Blumm and Barab{\'a}si, Albert-L{\'a}szl{\'o}},
  title        = {Limits of Predictability in Human Mobility},
  journal      = {Science},
  year         = {2010},
  volume       = {327},
  number       = {5968},
  pages        = {1018--1021},
  month        = feb,
  doi          = {10.1126/science.1177170},
}

@article{Rybski2013,
  author       = {Diego Rybski and Sergey V. Buldyrev and Shlomo Havlin and Fredrik Liljeros and HernÃ¡n A. Makse},
  title        = {{Scaling Laws of Human Broadcast Communication Enable Distinction between Human, Corporate and Robot Twitter Users}},
  journal      = {PLOS ONE},
  year         = {2013},
  volume       = {8},
  number       = {7},
  pages        = {e65774},
  doi          = {10.1371/journal.pone.0065774},
}

@inproceedings{Karagiannis2007,
  author       = {Thanasis Karagiannis and Jean-Yves Le Boudec and Mihailo Vojnovic},
  title        = {Power law and exponential decay of inter contact times between mobile devices},
  booktitle    = {Proceedings of the 13th Annual ACM International Conference on Mobile Computing and Networking (MobiCom '07)},
  year         = {2007},
  pages        = {183--194},
  doi          = {10.1145/1287853.1287875},
}

@article{Barnett2016,
  author       = {Ian Barnett and Jukka-Pekka Onnela},
  title        = {{Inferring Mobility Measures from GPS Traces with Missing Data}},
  journal      = {Biostatistics},
  year         = {2020},
  volume       = {21},
  number       = {2},
  pages        = {e98--e112},
  doi          = {10.1093/biostatistics/kxy059},
}

@article{Fawcett2006,
  author       = {Tom Fawcett},
  title        = {{An introduction to ROC analysis}},
  journal      = {Pattern Recognition Letters},
  year         = {2006},
  volume       = {27},
  number       = {8},
  pages        = {861--874},
  doi          = {10.1016/j.patrec.2005.10.010},
}

@article{Ruths2020,
  author       = {Derek Ruths and Caitrin Armstrong},
  title        = {{The Residence History Inference Problem}},
  journal      = {arXiv preprint arXiv:2003.04155},
  year         = {2020}
}

@article{Pappalardo2015,
  author       = {Pappalardo, L. and Simini, F. and Rinzivillo, S. and Pedreschi, D. and Giannotti, F.},
  title        = {Returners and explorers dichotomy in human mobility},
  journal      = {Nature Communications},
  volume       = {6},
  pages        = {8166},
  year         = {2015},
  doi          = {10.1038/ncomms9166}
}

@misc{nuts,
  title        = {{Nomenclature of Territorial Units for Statistics (NUTS)}},
  author       = {{European Commission, Eurostat}},
  howpublished = {\url{https://ec.europa.eu/eurostat/web/nuts}},
  note         = {Accessed: 2025-07-09},
  year         = {2025}
}

@misc{bayloc,
  author       = {Rosillo-Rodes, Pablo},
  title        = {\texttt{bayloc}},
  howpublished = {\url{https://github.com/pablorosillo/bayloc}},
  year         = {2025},
  note         = {{GitHub repository}}
}

@article{Barbosa2018,
  title={Human mobility: Models and applications},
  author={Barbosa, Hugo and Barthelemy, Marc and Ghoshal, Gourab and James, Charlotte R and Lenormand, Maxime and Louail, Thomas and Menezes, Ronaldo and Ramasco, Jos{\'e} J and Simini, Filippo and Tomasini, Marcello},
  journal={Physics Reports},
  volume={734},
  pages={1--74},
  year={2018},
  publisher={Elsevier}
}

@article{Song2010scaling,
  title={Modelling the scaling properties of human mobility},
  author={Song, Chaoming and Koren, Tal and Wang, Pu and Barab{\'a}si, Albert-L{\'a}szl{\'o}},
  journal={Nature Physics},
  volume={6},
  number={10},
  pages={818--823},
  year={2010},
  publisher={Nature Publishing Group}
}

@article{Gonzalez2008,
  title={Understanding individual human mobility patterns},
  author={Gonz{\'a}lez, Marta C and Hidalgo, C{\'e}sar A and Barab{\'a}si, Albert-L{\'a}szl{\'o}},
  journal={Nature},
  volume={453},
  pages={779--782},
  year={2008},
  publisher={Nature Publishing Group}
}

@inproceedings{Liu2016,
  title={Predicting the next location: A recurrent model with spatial and temporal contexts},
  author={Liu, Qiang and Wu, Shu and Wang, Liang and Tan, Tieniu},
  booktitle={Thirtieth AAAI conference on artificial intelligence},
  year={2016}
}

@inproceedings{Feng2018,
  title={Deepmove: Predicting human mobility with attentional recurrent networks},
  author={Feng, Jie and Li, Yong and Zhang, Chao and Sun, Funing and Meng, Fanchao and Guo, Ang and Jin, Depeng},
  booktitle={Proceedings of the 2018 World Wide Web Conference},
  pages={1459--1468},
  year={2018}
}

@article{Alessandretti2018,
  author  = {Alessandretti, Luca and Sapiezynski, Piotr and Sekara, Vedran and Lehmann, Sune and Baronchelli, Andrea},
  title   = {Evidence for a conserved quantity in human mobility},
  journal = {Nature Human Behaviour},
  volume  = {2},
  pages   = {485--491},
  year    = {2018},
  doi     = {10.1038/s41562-018-0364-x}
}

@article{Luca2021,
author = {Luca, Massimiliano and Barlacchi, Gianni and Lepri, Bruno and Pappalardo, Luca},
title = {A Survey on Deep Learning for Human Mobility},
year = {2021},
issue_date = {January 2023},
publisher = {Association for Computing Machinery},
address = {New York, NY, USA},
volume = {55},
number = {1},
journal = {ACM Comput. Surv.},
month = nov,
articleno = {7},
}

@article{Pappalardo2023,
  title={Future directions in human mobility science},
  author={Pappalardo, Luca and Manley, Ed and Sekara, Vedran and Alessandretti, Laura},
  journal={{Nature Computational Science}},
  volume={3},
  number={7},
  pages={588--600},
  year={2023},
  publisher={Nature Publishing Group US New York}
}

@article{pappalardo2021evaluation,
  title={Evaluation of home detection algorithms on mobile phone data using individual-level ground truth},
  author={Pappalardo, Luca and Ferres, Leo and Sacasa, Manuel and Cattuto, Ciro and Bravo, Loreto},
  journal={EPJ Data Science},
  volume={10},
  number={1},
  pages={29},
  year={2021},
  publisher={Springer Berlin Heidelberg}
}

@article{schlosser2021biases,
  title={Biases in human mobility data impact epidemic modeling},
  author={Schlosser, Frank and Sekara, Vedran and Brockmann, Dirk and Garcia-Herranz, Manuel},
  journal={arXiv preprint arXiv:2112.12521},
  year={2021}
}

@article{nijs2025data,
  title={Data bias in human mobility is a universal phenomenon but is highly location-specific},
  author={Nijs, Katinka den and Omodei, Elisa and Sekara, Vedran},
  journal={arXiv preprint arXiv:2508.00149},
  year={2025}
}

@article{gallotti2024distorted,
  title={Distorted insights from human mobility data},
  author={Gallotti, Riccardo and Maniscalco, Davide and Barthelemy, Marc and De Domenico, Manlio},
  journal={Communications Physics},
  volume={7},
  number={1},
  pages={421},
  year={2024},
  publisher={Nature Publishing Group UK London}
}

@article{Cabrera2025ASM,
  title={A systematic machine learning approach to measure and assess biases in mobile phone population data},
  author={Carmen Cabrera and Francisco Rowe},
  journal={ArXiv},
  year={2025},
  volume={abs/2509.02603}
}

@article{louail2014,
  title={From mobile phone data to the spatial structure of cities},
  author={Louail, Thomas and Lenormand, Maxime and Cantu Ros, Oliva G and Picornell, Miguel and Herranz, Ricardo and Frias-Martinez, Enrique and Ramasco, Jos{\'e} J and Barthelemy, Marc},
  journal={{Scientific Reports}},
  volume={4},
  pages={5276},
  year={2014},
  publisher={Nature Publishing Group UK London}
}

@article{bassolas2019,
  title={Hierarchical organization of urban mobility and its connection with city livability},
  author={Bassolas, Aleix and Barbosa-Filho, Hugo and Dickinson, Brian and Dotiwalla, Xerxes and Eastham, Paul and Gallotti, Riccardo and Ghoshal, Gourab and Gipson, Bryant and Hazarie, Surendra A and Kautz, Henry and others},
  journal={Nature Communications},
  volume={10},
  pages={4817},
  year={2019},
  publisher={Nature Publishing Group UK London}
}

@article{dong2024,
  title={Defining a city-delineating urban areas using cell-phone data},
  author={Dong, Lei and Duarte, Fabio and Duranton, Gilles and Santi, Paolo and Barthelemy, Marc and Batty, Michael and Bettencourt, Lu{\'\i}s and Goodchild, Michael and Hack, Gary and Liu, Yu and others},
  journal={{Nature Cities}},
  volume={1},
  number={2},
  pages={117--125},
  year={2024},
  publisher={Nature Publishing Group US New York}
}

@article{balcan2009,
author = {Duygu Balcan  and Vittoria Colizza  and Bruno GonÃ§alves  and Hao Hu  and JosÃ© J. Ramasco  and Alessandro Vespignani },
title = {Multiscale mobility networks and the spatial spreading of infectious diseases},
journal = {{Proceedings of the National Academy of Sciences USA}},
volume = {106},
number = {51},
pages = {21484-21489},
year = {2009},
doi = {10.1073/pnas.0906910106}
}

@article{bajardi2011,
  title={{Human mobility networks, travel restrictions, and the global spread of 2009 H1N1 pandemic}},
  author={Bajardi, Paolo and Poletto, Chiara and Ramasco, Jose J and Tizzoni, Michele and Colizza, Vittoria and Vespignani, Alessandro},
  journal={{PloS One}},
  volume={6},
  pages={e16591},
  year={2011},
  publisher={Public Library of Science San Francisco, USA}
}

@article{Abrar2023AnalysisOP,
  title={{Analysis of Performance Improvements and Bias Associated with the Use of Human Mobility Data in COVID-19 Case Prediction Models}},
  author={Saad Mohammad Abrar and Naman Awasthi and Daniel Smolyak and Vanessa Fr{\'i}as-Mart{\'i}nez},
  journal={ACM Journal on Computing and Sustainable Societies},
  year={2023},
  volume={1},
  pages={1 - 36},
}

@article{lu2026,
title = {Human mobility in epidemic modeling},
journal = {Physics Reports},
volume = {1157},
pages = {1-45},
year = {2026},
author = {Xin Lu and Jiawei Feng and Shengjie Lai and Petter Holme and Shuo Liu and Zhanwei Du and Xiaoqian Yuan and Siqing Wang and Yunxuan Li and Xiaoyu Zhang and Yuan Bai and Xiaojun Duan and Wenjun Mei and Hongjie Yu and Suoyi Tan and Fredrik Liljeros}
}

@article{barreras2024,
  title={{The exciting potential and daunting challenge of using GPS human-mobility data for epidemic modeling}},
  author={Barreras, Francisco and Watts, Duncan J},
  journal={{Nature Computational Science}},
  volume={4},
  number={6},
  pages={398--411},
  year={2024},
  publisher={Nature Publishing Group US New York}
}

@article{pullano2024mobility,
  title={Mobility resolution needed to inform predictive epidemic models for spatial transmission from mobile phone data},
  author={Pullano, Giulia and Bansal, Shweta and Rubrichi, Stefania and Colizza, Vittoria},
  journal={medRxiv},
  pages={2024--10},
  year={2024},
  publisher={Cold Spring Harbor Laboratory Press}
}
\end{document}